\documentclass{article}
\pdfoutput=1

\usepackage{microtype}
\usepackage{graphicx}
\usepackage{subfigure}
\usepackage{booktabs} 

\usepackage[accepted]{icml2024}

\usepackage{mathtools}
\usepackage{amsthm}
\PassOptionsToPackage{compress}{natbib}

\usepackage[utf8]{inputenc} 
\usepackage[T1]{fontenc}
\usepackage{url}
\usepackage{booktabs}    
\usepackage{amsfonts}     
\usepackage{nicefrac}       
\usepackage{microtype}      
\usepackage{xcolor}       
\usepackage[pagebackref,breaklinks,colorlinks,citecolor=blue]{hyperref}
\usepackage{amsmath}
\usepackage{multirow}
\usepackage{multicol}
\usepackage{amssymb}
\usepackage{longtable}
\usepackage{graphicx}
\usepackage{soul}
\usepackage[capitalize,noabbrev]{cleveref}

\theoremstyle{plain}
\newtheorem{theorem}{Theorem}[section]

\theoremstyle{definition}

\theoremstyle{remark}

\newcommand{\bl}[1]{{\color{blue}{#1}}}
\newcommand{\red}[1]{{\color{red}{#1}}}

\usepackage[textsize=tiny]{todonotes}

\icmltitlerunning{On the Duality Between Sharpness-Aware Minimization and Adversarial Training}

\begin{document}

\twocolumn[
\icmltitle{On the Duality Between\\Sharpness-Aware Minimization and Adversarial Training}

\icmlsetsymbol{corresponding}{${\!}\dagger$}
\icmlsetsymbol{equal}{*}

\begin{icmlauthorlist}
\icmlauthor{Yihao Zhang}{pku,equal}
\icmlauthor{Hangzhou He}{pku,equal}
\icmlauthor{Jingyu Zhu}{pku,ucb,equal}
\icmlauthor{Huanran Chen}{bit}
\icmlauthor{Yifei Wang}{mit}
\icmlauthor{Zeming Wei}{pku,corresponding}
\end{icmlauthorlist}

\icmlaffiliation{pku}{Peking University}
\icmlaffiliation{ucb}{University of California, Berkeley}
\icmlaffiliation{bit}{Beijing Institute of Technology}
\icmlaffiliation{mit}{MIT CSAIL}

\icmlcorrespondingauthor{Zeming Wei}{weizeming@stu.pku.edu.cn}

\icmlkeywords{Adversarial Robustness, Sharpness-Aware Minimization}

\vskip 0.3in
]



\printAffiliationsAndNotice{\icmlEqualContribution} 

\begin{abstract}
Adversarial Training (AT), which adversarially perturb the input samples during training, has been acknowledged as one of the most effective defenses against adversarial attacks, yet suffers from inevitably decreased clean accuracy. Instead of perturbing the samples, Sharpness-Aware Minimization (SAM) perturbs the model weights during training to find a more flat loss landscape and improve generalization. However, as SAM is designed for better clean accuracy, its effectiveness in enhancing adversarial robustness remains unexplored. In this work, considering the duality between SAM and AT, we investigate the adversarial robustness derived from SAM. Intriguingly, we find that using SAM alone can improve adversarial robustness. To understand this unexpected property of SAM, we first provide empirical and theoretical insights into how SAM can implicitly learn more robust features, and conduct comprehensive experiments to show that SAM can improve adversarial robustness notably without sacrificing any clean accuracy, shedding light on the potential of SAM to be a substitute for AT when accuracy comes at a higher priority. Code is available at \url{https://github.com/weizeming/SAM_AT}.
\end{abstract}

\section{Introduction}

The existence of adversarial examples~\citep{goodfellow2014explaining, szegedy2013intriguing} has raised serious safety concerns on the deployment of deep neural networks (DNNs). To defend against attacks of crafting adversarial examples, 
a variety of defense methods~\citep{papernot2016distillation,xie2019feature,cohen2019certified,chen2023robust} has been proposed. \textbf{Adversarial training (AT)}~\citep{madry2017towards}, which adds adversarial perturbations to the training samples in the training loop, has been acknowledged as one of the most effective adversarial defense paradigms~\citep{athalye2018obfuscated}. 
Despite this success, AT suffers from an intrinsic limitation that decreases the clean accuracy, leading to a fundamental trade-off between accuracy and robustness~\cite{tsipras2018robustness,zhang2019theoretically}. A critical drawback of AT is that the perturbation of training samples makes the training distribution deviate from the natural data distribution.

Instead, \textbf{Sharpness-Aware Minimization (SAM)}~\cite{foret2020sharpness} perturbs model weights yet keeps using the original samples during training. SAM is a novel training framework that regularizes the sharpness of the loss landscape by weight perturbation to improve generalization ability. So far, SAM has achieved remarkable success in modern machine learning~\cite{bahri2021sharpness,andriushchenko2022towards,chen2023rethinking} and many of its follow-up variants have been proposed~\cite{kwon2021asam,du2021efficient,chen2022bootstrap}. However, as SAM was originally designed for better natural generalization, its impact on adversarial robustness remains unexplored.

In this paper, considering the duality between SAM and AT, \textit{i.e.} \textbf{perturbing samples (AT)} and \textbf{perturbing weights (SAM)}, we explore whether SAM can also achieve better adversarial robustness. 
Surprisingly, we find that using SAM alone can notably improve adversarial robustness compared to standard optimization methods like Adam, which is an unexpected benefit of SAM. As illustrated in Table~\ref{tab:intro}, SAM can achieve both higher robust and clean accuracy than standard training. Compared to the improvement in clean accuracy, the enhancement of robust accuracy (evaluated by AutoAttack~\cite{croce2020reliable}) is significantly prominent (\textbf{from 3.4\% to 25.4\%}). By contrast, though AT can exhibit better robustness, their decrease in natural accuracy is unaffordable in practical deployment. Motivated by this intriguing observation, we raise the two following questions and attempt to answer them: 
\vspace{3pt}\\
\textbf{(1) Why can SAM improve adversarial robustness}, and \vspace{3pt}\\
\textbf{(2) Under what condition can SAM be deployed as a substitution for AT.} 

\begin{table}[t]
    \centering
    \caption{Examples of clean accuracy and robust accuracy with AutoAttack (AA.) comparison on standard training, SAM, and AT. The robustness is evaluated under $\ell_2$ norm $\epsilon=32/255$. More details in Section~\ref{Section 5}.}
    \begin{tabular}{c|cc}
    \toprule
    Method & Clean Acc. & AA. Rob. Acc. \\
    \midrule
    Adam & 76.0 & 3.4 \\
    SAM & \textbf{\bl{78.7}} & \textbf{25.4}  \\
    \midrule
    $\ell_\infty$-AT & \red{58.3} & 52.4 \\
    $\ell_2$-AT & \red{64.2} & 57.5 \\
     \bottomrule
    \end{tabular}
    \vspace{-10pt}
    \label{tab:intro}
\end{table}

To study the two questions above, we first provide an intuitive understanding of why SAM can improve adversarial robustness without sacrificing natural accuracy. Specifically, we show that weight perturbation during training can help the model implicitly learn the robust features~\cite{tsipras2018robustness}. To verify this notion, we also provide theoretical insights to support our intuitive understanding. Following the popular data distribution based on robust and non-robust features decomposition~\cite{tsipras2018robustness}, we show that both SAM and AT can improve the robustness of the trained models by \textbf{biasing more weight on robust features}. To sum up, we answer the raised question \textbf{(1)} by concluding that SAM can improve adversarial robustness by implicitly learning robust features.

Furthermore, we conduct extensive experiments across different tasks,  data modalities, and model architectures to evaluate the robustness improvement of using SAM, where we show that under various settings, SAM exhibits consistently better robustness than standard training methods while still maintaining higher natural performance. Besides, though AT can achieve higher robustness than SAM, it inevitably sacrifices natural performance which may be unaffordable for real-world applications. 
Therefore, we answer the raised question \textbf{(2)} with the conclusion that SAM can be used as a substitute for AT 
when accuracy is more important but better robustness is preferred.

To summarize, our main contributions in this paper are:
\begin{enumerate}
\item We uncover an intriguing property of SAM that it can notably enhance adversarial robustness while maintaining natural 
performance compared to standard training, which is an unexpected benefit.

\item  We provide empirical and theoretical insights to understand how SAM can enhance adversarial robustness by showing that both input and weight perturbation can encourage the model to learn robust features. 

\item We conduct extensive experiments to show the effectiveness of SAM in terms of enhancing robustness without sacrificing natural performance. We also suggest that SAM can be considered a lightweight substitute for AT when accuracy comes at a higher priority. 

\end{enumerate}

\section{Background and Related Work}
\label{Section 2}
\subsection{Sharpnes-Aware Minimization (SAM)}
To improve generalization ability in traditional machine learning algorithms, \cite{hochreiter1994simplifying,hochreiter1997flat} respectively attempt to search for flat minima and penalize sharpness in the loss landscape, which obtains good results in generalization~\cite{keskar2016large,neyshabur2017exploring,dziugaite2017computing}. 
Inspired by this, a series of works focus on using the concept of \textit{flatness} or \textit{sharpness} in loss landscape to ensure better generalization, \textit{e.g.} Entropy-SGD~\cite{chaudhari2019entropy} and Stochastic Weight Averaging (SWA)~\cite{izmailov2018averaging}.
Sharpness-Aware minimization (SAM)~\cite{foret2020sharpness} also falls into this category, which simultaneously minimizes loss value and loss sharpness as described in (\ref{eq1}). 

The objective of SAM is to minimize the \textit{sharpness} around the parameters, which can be formulated as 
\begin{equation}\label{eq1}
\min_w \mathbb E_{(x,y)\sim\mathcal D}   \max_{\|\epsilon\|\le\rho;x,y} L(w+\epsilon)+\lambda \|w\|_2^2,
\end{equation}
where $\mathcal D$ is the data distribution, $L$ is the loss function, $w$ is the parameters of the model,
$\|w\|_2^2$ is the regularization term and $\rho$ controls the magnitude of weight perturbation. To solve the inner maximization process, one-step gradient descent is applied.

So far, SAM has become a powerful tool for enhancing the natural accuracy performance of machine learning models.
There are also many applications of SAM in other fields of research like language models \cite{bahri2021sharpness} and decentralized SGD~\citep{zhu2023decentralized}, showing the scalability of SAM to various domains. In addition, many improvements of the algorithm SAM spring up, like Adaptive SAM (ASAM)~\cite{kwon2021asam}, Efficient SAM (ESAM) \cite{du2021efficient}, LookSAM~\cite{liu2022towards}, Sparse SAM (SSAM)~\cite{mi2022make}, Fisher SAM~\cite{kim2022fisher} and SAM-ON~\cite{mueller2023normalization}, which add some modifications on SAM and further improve the generalization ability of the model. However, while these existing works focus on the natural generalization goal, the effectiveness of SAM on adversarial robustness remains unexplored.

\subsection{Adversarial Robustness and Adversarial Training}
The adversarial robustness and adversarial training have become popular research topics since the discovery of adversarial examples~\cite{szegedy2013intriguing,goodfellow2014explaining}, which uncovers that DNNs can be easily fooled to make wrong decisions by adversarial examples that are crafted by adding small perturbations to normal examples. The malicious adversaries can conduct adversarial attacks by crafting adversarial examples, which cause serious safety concerns regarding the deployment of DNNs. So far, numerous defense approaches have been proposed~\cite{xie2019feature,bai2019hilbert,cohen2019certified,chen2024your}, among which Adversarial Training (\textbf{AT})~\cite{madry2017towards} has been considered the most promising defense method against adversarial attacks. AT can be formulated as the following optimization problem:
\begin{equation}
    \min_w \mathbb E_{(x,y)\sim \mathcal D}\max_{\|\delta\|\le\epsilon} L(w; x+\delta,y),
\end{equation}
where $\mathcal D$ is the data distribution, $\epsilon$ is the margin of perturbation, $w$ is the parameters of the model and $L$ is the loss function (\textit{e.g.} the cross-entropy loss).
For the inner maximization process, the Projected Gradient Descent (PGD)~\cite{madry2017towards} attack is commonly used to generate the adversarial example:
\begin{equation}
x^{t+1}=\Pi_{\mathcal B(x,\epsilon)} (x^t+\alpha\cdot\text{sign}(\nabla_{x} \ell(\theta;x^t,y))),
\label{PGD}
\end{equation}
where $\Pi$ projects the adversarial example onto the perturbation bound $\mathcal B(x,\epsilon) = \{x':\|x'-x\|_p\le \epsilon\}$ and $\alpha$ represents the step size of gradient ascent.

Though improves adversarial robustness effectively, adversarial training has exposed several defects such as 
computational overhead~\cite{shafahi2019adversarial}, class-wise fairness~\cite{xu2021robust,wei2023cfa} and robust overfitting~\cite{rice2020overfitting,wang2024balance,wei2023characterizing} among which the decreased natural accuracy~\cite{tsipras2018robustness,zhang2019theoretically} has become the major concern.  One explanation for this drawback of AT is perturbing the samples during training leads the training sample distribution to deviate from the natural data~\cite{Ilyas2019AdversarialEA}.

In the context of adversarial robustness, several works also attempt to introduce a flat loss landscape in adversarial training~\cite{wu2020adversarial,yu2022robust,yu2022robust2}. The most representative one is Adversarial Weight Perturbation (AWP)~\cite{wu2020adversarial}, which simultaneously adds perturbation on examples and feature space to apply SAM and AT.
However, AWP also suffers from a decrease in natural accuracy which is even lower than AT in some cases, which we assume is because perturbing both the inputs and parameters significantly raises the difficulty of robust learning. We compare SAM and AWP in our experiment section. Besides this thread of work, the adversarial robustness derived from SAM alone has not been explored.

\section{Empirical Understanding}
\label{Section 3}

In this section, we provide an intuitive explanation to empirically understand how SAM improves adversarial robustness by demonstrating the duality between SAM and AT. Specifically, considering the arithmetical duality of the input and parameters to get the output in a specific layer, we can assume that the robustness against weight perturbation may also lead to robustness against input perturbation.

We start by rewriting the optimization objective of SAM and AT in a unified form and omit the regularization term $\lambda\|w\|_2^2$ as follows:
 \begin{equation}
 \min_{\boldsymbol w}\mathbb E_{(x,y)\sim \mathcal D}\max_{||\boldsymbol{\epsilon}||<\rho} L(\boldsymbol{w}+\boldsymbol{\epsilon};x,y)\quad \text{(SAM)}
 \end{equation}
 and
 \begin{equation}
     \min_{\boldsymbol w} \mathbb E_{(x,y)\sim \mathcal D}\max_{\|\delta\|\le\epsilon} L({\boldsymbol w}; x+\delta,y)
     \quad \text{(AT)}
 \end{equation}
To illustrate their relation, we first emphasize that both techniques involve adding \textbf{perturbation} to make the output more robust \textit{w.r.t.} input or weight changes since they both utilize one or more step gradient optimization to solve the inner maximization problem. However, as AT \textbf{explicitly} adds these perturbations to input examples $x$ and transforms them into adversarial examples $x_{adv}$, the adversarial sample distribution learned through forward-backward passes deviated from the natural distribution, leading to an inevitable decrease when evaluating natural performance in the original distribution. By contrast, SAM applies weight perturbation to achieve this robustness but keeps the original samples during learning, which can implicitly bias more weight on robust features~\cite{Ilyas2019AdversarialEA}.%

To be more formalized, we illustrate our understanding with a middle linear layer in a model, which extracts feature $z$ from input $x$: $z = Wx$. During AT, we add perturbations directly to the input space, resulting in $x \gets x + \delta$. However, in SAM, the perturbation is not directly applied to the input space, but to the parameter space as $W \gets W + \delta$. This leads to $Wx + W\delta$ for input perturbation and $Wx + \delta x$ for parameter perturbation. If the weight $W$ can keep $Wx$ more robust against small perturbations around it in the subsequent layers, it will be also beneficial to improve sample robustness around $x$.

Besides, we discuss the attack (perturbation) strength of AT and SAM. For SAM, the perturbation is relatively more moderate, as its perturbations are conducted in the weight space and do not change the original input. On the other hand, in order to achieve the best robustness by eliminating the non-robust features~\cite{Ilyas2019AdversarialEA}, AT applies larger and more straightforward perturbations to the input space, leading to better robustness but a loss in natural accuracy, which is not the original goal of SAM, but it is the goal of AT.

In summary, our intuitive analysis suggests that SAM applies small perturbations implicitly to the feature space to maintain good natural accuracy performance, while AT utilizes direct input perturbations, which may result in a severe loss in natural accuracy. We provide more theoretical evidence to support these claims in the next section.

\section{Theoretical Insights}
\label{Section 4}

In this section, we provide a theoretical analysis of SAM and the relation between SAM and AT. Following the \textit{robust/non-robust feature} decomposition~\cite{tsipras2018robustness}, we introduce a simple binary classification model, in which we show the implicit similarity and differences between SAM and AT.
We first present the data distribution and hypothesis space, then present how SAM and AT work in this model respectively, and finally discuss their relations.

\subsection{A Binary Classification Model}
Following a series of theoretical work on adversarial robustness~\cite{tsipras2018robustness,Ilyas2019AdversarialEA,xu2021robust}, we consider a similar binary classification task that the input-label pair $(\boldsymbol x,y)$ is sampled from $\boldsymbol x\in\{-1,+1\}\times \mathbb R^{n+1}$ and $y\in\{-1,+1\}$, and the distribution $\mathcal{D}$ is defined as follows.
\begin{equation} 
\begin{aligned}
& y \stackrel{\text { u.a.r }}{\sim}\{-1,+1\}, x_{1}=\{\begin{array}{ll}
+y, & \text { w.p. } p, \\
-y, & \text { w.p. } 1-p,
\end{array} \\ &  x_{2}, \ldots, x_{n+1} \stackrel{i . i . d}{\sim} \mathcal{N}(\eta y, 1),
\end{aligned}
\end{equation}
where $p\in(0.5,1)$ is the accuracy of feature $x_1$, constant $\eta>0$ is a small positive number.
In this model, $x_1$ is called the \textit{robust feature}, since any small perturbation can not change its sign. 
However, the robust feature is not perfect since $p<1$. 
Correspondingly, the features $x_2,\cdots,x_{n+1}$ are useful for identifying $y$ due to the consistency of sign,
hence they can help classification in terms of natural accuracy.
However, they can be easily perturbed to the contrary side (change their sign) since $\eta$ is a small positive, which makes them called \textit{non-robust features}~\cite{Ilyas2019AdversarialEA}.

Now consider a linear classifier model which predicts the label of a data point by computing $f_{\boldsymbol w}(\boldsymbol x) = \mathrm{sgn}(\boldsymbol w \cdot \boldsymbol x)$, and optimize the parameters $w_1,w_2,\cdots,w_{n+1}$ to maximize $\mathbb{E}_{x.y \sim \mathcal{D}} \mathbf{1}(f_{\boldsymbol w}(x) = y).$ 
Note that value $w_1$ as the coefficient of the robust feature $x_1$, may have a strong correlation with the robustness of the model.
Specifically, larger $w_1$ indicates that the model biases more weight on the robust feature $x_1$ and less weight on the non-robust features $x_2,\cdots, x_{n+1}$, leading to better robustness. {Therefore, we consider the weight of parameter $w_1$ among all features as an indication of how many robust features are learned in the model. Therefore, we define the \textbf{robust feature weight} $W_R$ of a given model as
  \begin{equation}
 W_R = \frac{w_1}{w_2+w_3+\cdots +w_{n+1}}
  \end{equation}
  to \textbf{measure the weight of robust features involved} in the model prediction.  We use the following lemma to justify the fundamental relationship between $W_R$ and the adversarial robustness of the model:

\begin{theorem}
The robust accuracy ($R_A$) of this model, defined as
\begin{equation}
R_A=\mathbb{E}_{\mathbf{x},y\sim D}\underset{||\delta||<\epsilon}{\min}\mathbf{1}\{{f_{\boldsymbol{w}}(x+\delta)=y}\}.
\end{equation}
is a \textbf{monotonic increasing function} of $W_R$ under condition $\epsilon<\eta$ and $0<W_R<W_R^{AT}$(defined in (\ref{AT})).
\end{theorem}

In the following, we derive and compare the robust feature weight $W_R$ of the trained model under standard training (ST), AT, and SAM respectively. To make our description clear, we denote the loss function $\mathcal{L}(\boldsymbol x,y,w)$ as $1 - \mathrm{Pr}(f_w(\boldsymbol x) = y)$ and for a given $\epsilon>0$, we define the loss function of SAM $\mathcal{L}^{SAM} $ as $ \max_{|\delta| \le\epsilon}\mathcal{L}(\boldsymbol x,y,w+\delta). $

\subsection{Standard Training (ST)}
We first show that under standard training, the robust feature weight learned in this model can be derived from the following theorem:
\begin{theorem}[Standard training]
    \label{theorem ST}
    In the model above, under standard training, the robust feature weight $W_R$ is 
    \begin{equation}
    W_R ^{*}= \frac{\ln p - \ln (1-p)}{2n\eta}.    
    \end{equation}
\end{theorem}
Therefore, $W_R^*$ can be regarded as the measurement $W_R$ returned by standard training with this model.

\subsection{Adversarial Training (AT)}
Now we consider when AT is applied to train the model. Recall that in this case, the loss function is no longer the standard one but the expected adversarial loss 
\begin{equation}
    \underset{(\boldsymbol x, y) \sim \mathcal{D}}{\mathbb{E}}\left[\max _{||\boldsymbol\delta||_{\infty} \le \epsilon} \mathcal{L}(\boldsymbol x+\boldsymbol\delta, y ; w)\right].
\end{equation}
Similar to standard training, we can derive the robust feature weight from the following theorem:
\begin{theorem}[Adversarial training]
\label{theorem AT}
In the classification problem above, under adversarial training with perturbation bound $\epsilon < \eta$, 
the robust feature weight 
\begin{equation}
\label{AT}
  W_R^{AT} = \frac{\ln p - \ln (1-p)}{2n(\eta-\epsilon)}.  
\end{equation}
\end{theorem}
We can see that {$W_R$} has been multiplied by $\frac{\eta}{\eta-\epsilon} > 1,$ which has \textbf{increased} the dependence on the robust feature $x_1$ of the classifier. 
This shows the adversarially trained model pays more attention to robust features compared to the standard-trained one, which improves adversarial robustness.

\subsection{Sharpness-Aware Minimization (SAM)}
Now we consider the situation of SAM. Recall that the optimizing objective of SAM is
\begin{equation}
    \underset{(x, y) \sim \mathcal{D}}{\mathbb{E}}\left[\max _{|\delta| \le \epsilon} \mathcal{L}(x, y ; w+\delta)\right].
\end{equation}
 We first explain why SAM could improve the adversarial robustness by proving that the measurement $W_R$ trained with SAM $W_R^{SAM}$ is also larger than $W_R^*$, which is stated as follows:
\begin{theorem}[Sharpness-aware minimization]
    \label{SAM1}
    In the classification problem above, the robust feature weight for SAM training {$W_R^{SAM}$} satisfies that
    \begin{equation}
        W_R^{SAM}> W_R^*.
    \end{equation}
\end{theorem}
From Theorem \ref{theorem AT} and \ref{SAM1} {we can see that both $W_R^{AT}$ and $W_R^{SAM}$ are greater than $W_R^{*}$}, which indicates both SAM and AT encourages the trained model to learn more robust features.
However, the qualitative relation is not sufficient to quantify how much robustness SAM achieves compared to adversarial training, and we attempt to step further by quantitatively estimating the {$W_R^{SAM}$} in the following theorem:
\begin{theorem}
    \label{SAM2}
    In the classification problem above, suppose that $\epsilon>0$ is small, we have \begin{equation}{W_R^{SAM}\approx W_R^{*}+\frac{2}{3}W_R^*\epsilon^{2}}.
    \end{equation}
\end{theorem}

\subsection{Relation between SAM and AT}

We further discuss the distinct attack (perturbation) strength between AT and SAM.
Recall that in our empirical understanding in  Section~\ref{Section 3}, the perturbation of SAM is more moderate and implicit than AT.
Therefore, to reach the same robustness level (which is measured by the robust feature weight $W_R$), SAM requires a much larger perturbation range, while for AT, less perturbation over $x$ is enough.
Theoretically, the following theorem verifies our statement:
\begin{theorem} 
\label{relation}
Denote the perturbation range $\epsilon$ of AT and SAM  as $\epsilon_{AT}$ and $\epsilon_{SAM}$, respectively. Then, when both methods return the same robust feature weight $W_R$, we have the following relation between $\epsilon_{AT}$ and $\epsilon_{SAM}$:
\begin{equation}
2+\frac{3}{\epsilon_{SAM}^2}\approx\frac{2\eta}{\epsilon_{AT}}
\label{eq32}
\end{equation}
\end{theorem}
From theorem~\ref{relation}, we can identify the different perturbation strengths of AT and SAM.
It can be easily derive from Theorem \ref{relation} that $\epsilon_{SAM}$ is larger than $\epsilon_{AT}$ when (\ref{eq32}) holds, since we assume $\eta$ is a small positive, $\epsilon$ is small in Theorem~\ref{SAM2} and $\epsilon_{AT}<\eta$ in Theorem~\ref{theorem AT}.
Therefore, to gain the same weight $w_1$ on robust features $x_1$, $\epsilon_{AT}$ only need to be chosen much smaller than $\epsilon_{SAM}$. On the other hand, under the same perturbation bound  $\epsilon_{AT}=\epsilon_{SAM}$, the model trained under AT has larger $W_R$ than SAM, hence it focuses on more robustness yet decreases more natural accuracy.

All proofs can be found in Appendix~\ref{proofs}. While we acknowledge that $W_R$ is only analyzed under a simple model and cannot be directly generalized to multiple-layer networks, we believe the insights delivered from $W_R$ can be generalized to DNNs. To sum up, we can conclude that AT utilizes explicit and direct perturbations for eliminating non-robust features, while SAM leverages implicit and moderate perturbations to learn robust features. This is consistent with our empirical understanding in Section~\ref{Section 3} and we also verify these claims with experiments in the following section.

\section{Experiment}
\label{Section 5}

\begin{table*}[t]
\caption{Natural and robust accuracy evaluation on \textbf{CIFAR-10} dataset.}
    \centering
    \begin{tabular}{c|c|ccccccc|ccccc}  
    \toprule

        \multirow{2}{*}{Method}
        &  
        {Natural}
       & FGSM & $\ell_\infty$-PGD & $\ell_2$-PGD   
       & $\ell_2$-AA. 
       & \multirow{2}{*}{StAdv} & \multirow{2}{*}{FAB} & \multirow{2}{*}{Pixle} & Average
       \\

       & Accuracy & $\epsilon=\frac{1}{255}$ &  $\epsilon=\frac{1}{255}$ & $\epsilon=\frac{32}{255}$ &  $\epsilon=\frac{32}{255}$ &  & & & Robustness\\
       \midrule
        SGD & 94.5 & 63.4 & 37.9 & 41.5 & 31.7 & 35.2 & 44.8 & 10.0 & \textbf{37.8} \\
        Adam & 93.9 & 44.3 & 17.4 & 20.7 & 13.9 & 20.4 & 24.7 & 7.6 & 21.3 \\
        \midrule
        SAM ($\rho=0.1$) & 95.4 & 63.3 & 46.2 & 48.7 & 43.6 & 39.3 & 49.2 & 13.4 & 43.4 \\
        SAM ($\rho=0.2$) & \textbf{95.5} & 66.7 & 51.3 & 53.4 & 48.1 & 44.2 & 53.4 & 13.2 & 47.2 \\
        SAM ($\rho=0.3$) & 95.4 & 66.6 & 51.2 & 53.5 & 47.8 & 46.1 & 53.8 & 13.7 & 47.5 \\
        SAM ($\rho=0.4$) & \bl{94.7} & 69.6 & 56.4 & 58.6 & 51.8 & 54.9 & 57.6 & 14.3 & \textbf{51.9} \\
        \midrule
        AT ($\ell_\infty$-$\epsilon=\frac{8}{255}$) & \red{84.5} & 81.9 & 81.8 & 79.7 & 79.5 & 82.0 & 79.5 & 26.9 & 73.0 \\
        \midrule
        AT ($\ell_2$-$\epsilon=\frac{128}{255}$) & \red{89.2} & 84.1 & 84.1 & 84.8 & 84.8 & 80.4 & 84.8 & 32.0 & 76.4 \\
        \bottomrule

    \end{tabular}
\vspace{-10pt}
    \label{tab:exp:classification_cifar10}
\end{table*}

\begin{table*}[t]
\caption{Natural and robust accuracy evaluation on \textbf{CIFAR-100} dataset.}
    \centering
    \begin{tabular}{c|c|ccccccc|ccccc}  
    \toprule

        \multirow{2}{*}{Method}
        &  
        {Natural}
       & FGSM & $\ell_\infty$-PGD & $\ell_2$-PGD   
       & $\ell_2$-AA. 
       & \multirow{2}{*}{StAdv} & \multirow{2}{*}{FAB} & \multirow{2}{*}{Pixle} & Average
       \\
       
       & Accuracy & $\epsilon=\frac{1}{255}$ &  $\epsilon=\frac{1}{255}$ & $\epsilon=\frac{32}{255}$ &  $\epsilon=\frac{32}{255}$ &  & & & Robustness\\
       \midrule
        SGD & 76.5 & 30.9 & 13.3 & 17.0 & 11.4 & 14.5 & 16.9 & 1.8 & \textbf{15.1} \\
        Adam & 76.0 & 20.5 & 5.3 & 5.8 & 3.4 & 7.4 & 8.4 & 1.3 & 7.4 \\
        \midrule
        SAM ($\rho=0.1$) & 77.7 & 36.3 & 20.3 & 24.1 & 19.6 & 20.9 & 24.5 & 3.4 & 21.3 \\
        SAM ($\rho=0.2$) & \textbf{78.8} & 38.1 & 22.9 & 25.6 & 20.2 & 23.7 & 25.2 & 3.7 & 22.8 \\
        SAM ($\rho=0.3$) & 78.7 & 40.2 & 25.5 & 28.1 & 22.6 & 26.4 & 27.5 & 4.3 & 24.9 \\
        SAM ($\rho=0.4$) & \bl{78.7} & 41.8 & 29.5 & 31.8 & 25.4 & 29.4 & 30.0 & 5.4 & \textbf{27.6} \\
        \midrule
        AT ($\ell_\infty$-$\epsilon=\frac{8}{255}$) & \red{58.3} & 55.1 & 55.0 & 53.8 & 52.4 & 50.6 & 52.7 & 7.3 & 46.7 \\
        \midrule
        AT ($\ell_2$-$\epsilon=\frac{128}{255}$) & \red{64.2} & 56.6 & 56.6 & 58.1 & 57.5 & 52.2 & 58.2 & 10.5 & 49.9 \\
        \bottomrule

    \end{tabular}
\vspace{-10pt}
    \label{tab:exp:classification_cifar100}
\end{table*}

In this section, we conduct extensive experiments to show the effectiveness of SAM in improving robustness while maintaining natural performance, across multiple tasks, data modalities, and various settings. We start with the classic image classification task, seconded by semantic segmentation (vision) and text classification (language).

\subsection{Image Classification}

\subsubsection{Experimental settings}

\textbf{Training configurations.} We mainly consider the vanilla SAM~\cite{foret2020sharpness} optimizer with the perturbation hyper-parameter $\rho$ from the range $\{0.1,0.2,0.3,0.4\}$. In addition, we also explore two distinct variants of SAM including Adaptive Sharpness-Aware Minimization (ASAM)~\citep{kwon2021asam} and Efficient Sharpness-aware Minimization (ESAM)~\citep{du2022efficient}. Following~\citet{pang2020bag} and~\citet{wei2023cfa}, we set the weight decay as \texttt{5e-4} and momentum as 0.9 and train 100 epochs with the learning rate initialized as 0.1 for SGD and \texttt{1e-3} for Adam, and is divided by 10 at the 75th and 90th epochs, respectively. For AT, we consider both $\ell_2$ and $\ell_\infty$ norms.

\textbf{Datasets and models. }
We examine the robustness of SAM on CIFAR-\{10,100\}~\cite{krizhevsky2009learning} and TinyImageNet~\citep{chrabaszcz2017downsampled} datasets. We mainly conduct our experiment with PreActResNet-18 (PRN-18)~\cite{he2016identity}. To demonstrate the scalability of SAM, we also include the Wider ResNet (WRN-28-10)~\cite{DBLP:journals/corr/ZagoruykoK16} and the vision transformer architecture (DeiT)\cite{touvron2021training}. 
To evaluate the corruption robustness of various models, we also use the CIFAR-10C dataset~\cite{hendrycks2019benchmarking}, a common corruptions dataset for CIFAR10 with 1000 randomly selected examples from CIFAR-10C and level-3 of perturbation severity for each corruption types (fog, snow, \textit{etc}).

\textbf{Attacks for robustness evaluations. }
We assess the model's resistance through a multifaceted approach to thoroughly explore its robustness against different threat models. We apply the Fast Gradient Sign Method (\textbf{FGSM})~\cite{goodfellow2015explaining} with a perturbation bound of $\ell_\infty$-$\epsilon=1/255$, and a 10-step Projected Gradient Descent (\textbf{PGD}) attack under $\ell_\infty$ and $\ell_2$ norms with bounds of $\epsilon=1/255$ and $32/255$, respectively. 
Moreover, we incorporate the \textbf{AutoAttack} (AA.)~\cite{croce2021mind} to ensure a reliable evaluation.
Besides, we also consider other popular attacks like StAdv~\cite{xiao2018spatially}, FAB~\cite{croce2020minimally} (10 steps under $\ell_2$-$\epsilon=32/255$).
For black-box assessment, we utilize Pixle~\cite{Pomponi_2022}, a strong pixel-rearranging black-box attack, constrained to a maximum of five iterations. 
The attacks are mostly implemented by the \texttt{torchattacks}~\cite{kim2020torchattacks} framework to ensure a reliable assessment. For corruption robustness~\cite{hendrycks2019benchmarking}, which measures the robustness of DNNs against a wide range of real-world disturbances like noise, blur, or weather variations, we evaluate models on RobustBench~\cite{croce2021robustbench}, a standardized benchmark designed to evaluate the general robustness. We ran all experiments three times independently to report the average result and omitted the standard deviations since they are small (less than 0.5\%) and do not affect our claims.

\subsubsection{Experimental results}

\textbf{Comparison with standard training. }
As shown in \cref{tab:exp:classification_cifar10,tab:exp:classification_cifar100,tab:exp:classification_tinyimagenet}, all models trained with SAM exhibit significantly better natural accuracy and robustness compared to those trained with standard training (ST). In particular, higher robustness is achieved by using larger values of $\rho$ with SAM. Taking the CIFAR-100 dataset as an example, the model trained with $\rho=0.4$ demonstrates even multiple robust accuracy than ST, and its natural accuracy is still higher than that of ST.
Compared to the improvement in natural accuracy (approximately 2\%), the increase of robustness is more significant (\textbf{more than 10\% in average}). Moreover, on large datasets like Tiny-ImageNet, SAM still surpasses ST by a large margin.
Therefore, we conclude that SAM with a relatively larger weight perturbation bound $\rho$ is a promising technique for enhancing adversarial robustness without sacrificing natural accuracy. Notably, our results are consistent with concurrent work showing that the robustness of SGD is better than Adam~\cite{ma2023understanding}.
\begin{table}[t]
    \centering
    \caption{Comparison on Tiny ImageNet.}
    \begin{tabular}{c|c|cc}
    \toprule
    Config & Natural & AA. & StAdv \\
    \midrule
    SGD & 57.4 & 2.5 & 3.4
    \\
    SAM ($\rho = 0.4$) & \textbf{57.9} & 10.3 & \textbf{13.3}
    \\
       SAM ($\rho = 0.5$) & \bl{57.7} & \textbf{10.4} & 12.7
    \\
    \midrule
         $\ell_\infty$-AT($\epsilon=8/255$) & \red{32.3} & 27.7 & 25.1
    \\
         $\ell_2$-AT($\epsilon=128/255$) & \red{41.6} & 32.8 & 30.2
        \\
    \bottomrule
    \end{tabular}
    \vspace{-10pt}
    \label{tab:exp:classification_tinyimagenet}
\end{table}

\begin{table}[h]
    \centering
    \caption{Comparing SAM and AT with small $\epsilon$ on CIFAR-100 dataset.}
    \begin{tabular}{c|c|c|cc}
    \toprule
    Method & Config & Natural & AA. & Pixle \\
    \midrule
    \multirow{2}{*}{Standard}
    &
    SGD & 76.5 & 11.4 & 1.8
    
    \\
    & Adam & 76.0 & 3.4 & 1.3
    \\
    \midrule
    \multirow{2}{*}{SAM}
    & $\rho=0.4$ & \textbf{78.7} & 25.4 & 5.4\\
    & $\rho=0.5$ & \bl{77.8} & \textbf{30.1} & \textbf{8.1}\\
    \midrule
    \multirow{4}{*}{$\ell_\infty$-AT} & $\epsilon=1/255$ & \red{72.1} & 58.6 & 4.7 \\
    & $\epsilon=2/255$ & 69.4 & 56.8 & 4.2\\
    & $\epsilon=4/255$ & 64.3 & 55.7 & 6.3 \\
    & $\epsilon=8/255$ & 58.3 & 52.4 & 7.3 \\
    \midrule
    \multirow{4}{*}{$\ell_2$-AT} & $\epsilon=16/255$ & \red{74.3} & 57.9 & 5.7 \\
     & $\epsilon=32/255$ & 72.5 & 57.2 & 5.2  \\
     & $\epsilon=64/255$ & 69.0 & 57.3 & 8.2 \\
     & $\epsilon=128/255$ & 64.2 & 57.5 & 10.5 \\
     \bottomrule
    \end{tabular}
    \vspace{-10pt}
    \label{tab:cifar_at_sam}
\end{table}

\textbf{Comparison with AT. }
Regarding adversarially trained models, 
although there remains a large gap between the robustness obtained by SAM and AT, all adversarially trained models exhibit significant lower natural accuracy than standard training and SAM.
Particularly, as demonstrated in \cref{tab:cifar_at_sam,tab:exp:classification_backbone}, even training with extremely small perturbation bound like $\epsilon=1/255$ decreases natural accuracy at 3.8\% for CIFAR-100 and 0.9\% for CIFAR-10 datasets, respectively.
The larger perturbation bound $\epsilon$ used in AT, the worse natural accuracy is obtained by the corresponding model. Therefore, a key benefit of using SAM instead of AT is that there is no decrease in clean accuracy.
Additionally, note that AT requires significant computational overhead. Specifically, for 10-step PGD, all AT experiments require 10 times more computational cost compared to ST. Moreover, the cutting-edge efficient SAMs~\citep{liu2022towards,chen2022bootstrap} are even faster than FastAT~\citep{wong2020fast}.

\textbf{Scalability to various architectures. }
\cref{tab:exp:classification_backbone} presents the results for three distinct models: PRN-18, WRN, and DeiT. Regardless of the architecture, SAM consistently improved robustness against natural, adversarial, and StAdv attacks compared to standard SGD. On PRN-18, SAM yielded impressive gains on both AA and StAdv attacks, boosting robustness by 11.9\% and 4.1\%, respectively. Similar trends were observed with WRN, achieving robustness improvements of 2.4\% and 4.7\% for AA and StAdv attacks. Notably, even for DeiT, where SGD performance was lower, SAM still offered significant gains in robustness. These results convincingly demonstrate the broad applicability and scalability of SAM in enhancing robustness across diverse model architectures.

\begin{table}[h]
    \centering
    \caption{Comparison of different model architectures on CIFAR-10 dataset with $\epsilon = 1/255$.}
    \begin{tabular}{cc|c|ccc}
    \toprule
    Model & Method & Natural & AA. & StAdv \\
    \midrule
    \multirow{3}{*}{PRN-18}
    &
    SGD & 94.5 & 31.7 & 35.2
    \\
      & SAM & \textbf{95.4} & 43.6 & 39.3
    \\
        & $\ell_\infty$-AT & 84.5 & 81.4 & 82.0
    \\
        
    \midrule
    \multirow{3}{*}{WRN}    &
    SGD & 95.2 & 39.6 & 38.3
    \\
      & SAM & \textbf{95.6} & 42.0 & 43.0
    \\
        & $\ell_\infty$-AT & 87.2 & 79.7 & 76.6
    \\
        
    \midrule
    \multirow{3}{*}{DeiT} & 
    SGD & 69.2 & 21.3 & 11.2
    \\
      & SAM & \textbf{69.4} & 24.2 & 19.6
    \\
        & $\ell_\infty$-AT & 62.2 & 51.9 & 54.2
    \\
        
     \bottomrule
    \end{tabular}
    \vspace{-10pt}
    \label{tab:exp:classification_backbone}
\end{table}

\textbf{Variants of SAM. } 
As shown in \cref{tab:exp:different_sam}, ASAM and ESAM generally outperform standard optimizers (SGD and Adam) in terms of robustness, demonstrating the effectiveness of adaptive momentum for adversarial defense. Notably, ESAM achieved the highest robustness against StAdv attacks (49.6\% at $\rho=0.4$), while SAM yielded the best robustness against the AA attack (58.4\% at $\rho=0.5$). These results highlight the robustness of various SAMs against adversarial threats.
\begin{table}[h]
    \centering
    \caption{Comparison of variants of SAM on CIFAR-10 dataset.}
    \begin{tabular}{c|c|c|cc}
    \toprule
    Method & Config & Natural & AA. & StAdv \\
    \midrule
    \multirow{2}{*}{Standard}
    &
    SGD & 94.5 & 31.7 & 35.2
    
    \\
    & Adam & 93.9 & 13.9 & 20.4
    \\
    \midrule
    \multirow{3}{*}{SAM}
    & $\rho=0.3$ & \textbf{95.4} & 57.8 & 46.1\\
    & $\rho=0.4$ & 94.7 & 51.8 & 54.9\\
    & $\rho=0.5$ & {94.5} & \textbf{58.4} & \textbf{55.8}\\
    \midrule
    \multirow{3}{*}{ESAM} 
    & $\rho=0.3$ & \textbf{95.2} & 53.1 & 48.2\\
    & $\rho=0.4$ & 94.7 & 51.6 & \textbf{49.6}\\
    & $\rho=0.5$ & 94.4 & \textbf{53.6} & 49.3\\
    \midrule
    \multirow{3}{*}{ASAM} 
    & $\rho=0.3$ & 95.4 & 39.9 & 34.7 \\
    & $\rho=0.4$ & 95.5 & 40.4 & 35.6 \\
    & $\rho=0.5$ & \textbf{95.7} & \textbf{44.2} & \textbf{36.6} \\
     \bottomrule
    \end{tabular}
    \vspace{-10pt}
    \label{tab:exp:different_sam}
\end{table}

\textbf{Corruption Robustness. }
Unlike robustness against adversarial attacks, common corruption robustness captures a wider spectrum of realistic disturbances beyond artificially crafted perturbations, showcasing how models can handle varied and unexpected environmental changes. We evaluated SAM and AT on corrupted images in CIFAR-10C and calculated the average accuracy across 9 different common corruptions. We present these experimental results in~\cref{tab:exp:corruption_cifar10_avg}, with detailed accuracy for each corruption type available in~\cref{tab:exp:corruption_cifar10} in Appendix~\ref{app 2}.

Unlike in adversarial attack scenarios where SAM cannot outperform AT in terms of robustness, models trained with SAM have demonstrated significantly superior robustness compared to AT. As shown in the table, $\ell_p$-norm-AT performs even worse than standard training (15.67\% for $\ell_\infty$-AT with $\epsilon = 8/255$, compared to 34.91\% for SGD). In contrast, SAM achieves similar results on corrupted images without significant performance degradation, showcasing the stability of SAM in realistic perturbation scenarios. Surprisingly, we found that variants of SAM can perform even better on corrupted images compared to standard training (with an accuracy increase of around 2\%), highlighting the potential of the SAM method for addressing real-world robustness issues.	

\begin{table}
\caption{Accuracy on common corruptions from CIFAR-10C.} 
    \centering
    \begin{tabular}{c|c|c}  
    \toprule
        Method & Natural & Corruption (Avg.) \\
       \midrule
        SGD & 94.5 & 34.91 \\
        Adam & 93.9 & 29.60 \\
        \midrule
        SAM ($\rho=0.3$) & \textbf{95.4} & 32.60 \\
        ASAM ($\rho=0.3$) & \textbf{95.4} & \textbf{36.69} \\
        ESAM ($\rho=0.3$) & 95.2 & 36.29 \\
        \midrule
        $\ell_\infty$-AT($\epsilon=8/255$) & 84.5 & 15.67 \\
        $\ell_2$-AT($\epsilon=128/255$) & 89.2 & 23.13 \\
        \bottomrule
    \end{tabular}
    \vspace{-10pt}
    \label{tab:exp:corruption_cifar10_avg}
\end{table}

\textbf{Comparison with AWP. } 
Adversarial Weight Perturbation (AWP)~\cite{wu2020adversarial} is an adversarial training method designed for better adversarial robustness by introducing a double-perturbation mechanism that adversarially perturbs both inputs and weights during training. Specifically, AWP can be regarded as a combination of SAM and AT. As AWP also leverages the weight perturbation paradigm, we provide a comparison with AWP and SAM to showcase that SAM still outperforms AWP in terms of natural accuracy. 

The experimental results are shown in~\cref{tab:comparison_awp}. Similar to the comparison between SAM and AT, we can see that although AWP outperforms both SAM and AT in terms of adversarial robustness, the natural accuracy with SAM is still higher than AWP even though weight perturbation was incorporated in it, indicating the importance of using natural data instead of adversarial examples for training to maintain the natural accuracy.

\begin{table}[h]
\centering
\vspace{-10pt}
\caption{Comparison of SAM, AT, and AWP on CIFAR-10.}
\label{tab:comparison_awp}
\begin{tabular}{c|c|c}
\toprule
Method & Natural & AA.  \\ \midrule
SAM ($\rho=0.1$) & \textbf{95.4}                       & 43.6\\ 
SAM ($\rho=0.4$) & 94.7                       & 51.8 \\ \midrule
$\ell_\infty$-AT ($\epsilon=8/255$) & 84.5 & 79.5 \\ 
$\ell_\infty$-AWP ($\epsilon=8/255$) & {82.0} & 80.1\\ \midrule
$\ell_2$-AT ($\epsilon=128/255$)    & 89.2 & {84.8}\\ 
$\ell_2$-AWP ($\epsilon=128/255$)   & {89.7} & \textbf{86.3}\\ \bottomrule
\end{tabular}
\vspace{-10pt}
\end{table}

\subsection{Semantic Segmentation}

\textbf{Experimental settings.}
We train DeepLabv3~\cite{chen2017rethinking} with randomly initialized ResNet-50~\cite{he2016deep} backbone on the Stanford background dataset~\cite{background_dataset} with Cross-Entropy loss for 30000 iterations. For experiments on VOC2012 dataset~\cite{voc-2012}, we use mobilenetv2~\cite{mobilenetv2} backbone, and the weights are initialized with pre-trained on ImageNet. The learning rate is initialized as 0.01 with a polynomial decay scheme. For the optimizer, the weight decay is set to \texttt{1e-4}, and the momentum is set to 0.9. We use the \textbf{mean Intersection over Union (mIoU)}~\cite{mIoU} to evaluate the segmentation results, where the IoU is calculated for each class at the pixel level as
\begin{equation}
{\rm IoU} = \frac{{\rm TP}}{{\rm TP}+{\rm FN}+{\rm FP}}\times 100\%,
\label{IoU}
\end{equation}
where TP, FN and FP represent true positive, false negative and false positive, respectively. The mIoU is then the mean value across all the classes in the dataset. For SAM, we select $\rho$ from $\{0.01, 0.02, 0.03, 0.04, 0.05\}$. 
\begin{table}[h]
    \centering
    \vspace{-10pt}
    \caption{Comparing SAM and AT for semantic segmentation on Stanford background dataset, mIoU is used for evaluation.}
    \begin{tabular}{c|c|c|cc}
    \toprule
    Method & Config & Natural & $\ell_\infty$-PGD & $\ell_2$-PGD \\
    \midrule
    \multirow{2}{*}{Standard}
    &
    SGD & 64.0 & 57.3 & 57.0
    
    \\
    & Adam & 62.2 & 57.1 & 56.8
    \\
    \midrule
    \multirow{5}{*}{SAM}
    & $\rho=0.01$ & 64.3 & 58.2 & 58.0\\
    & $\rho=0.02$ & \textbf{64.8} & \textbf{58.7} & 57.6\\
    & $\rho=0.03$ & 64.5 & 58.3 & \textbf{58.0}\\
    & $\rho=0.04$ & 64.6 & 57.9 & 57.6\\
    & $\rho=0.05$ & \bl{64.6} & 57.8 & 57.4\\
    \midrule
    \multirow{4}{*}{$\ell_\infty$-AT} & $\epsilon=1/255$ & \red{59.9} & \textbf{59.0} & \textbf{59.2}\\
    & $\epsilon=2/255$ & 58.2 & 57.7 & 57.8\\
    & $\epsilon=4/255$ & 55.3 & 54.9 & 55.1\\
    & $\epsilon=8/255$ & 57.5 & 55.1 & 54.9\\
     \bottomrule
    \end{tabular}
    \vspace{-10pt}
    \label{tabel-seg}
\end{table}
\begin{table}[h]
    \centering
    \caption{Comparing SAM and AT for semantic segmentation on VOC2012 dataset, mIoU on the validation set is used for evaluation.}
    \begin{tabular}{c|c|c|cc}
    \toprule
    Method & Config & Natural & $\ell_\infty$-PGD & $\ell_2$-PGD \\
    \midrule
    \multirow{2}{*}{Standard}
    &
    SGD & 66.9 & 37.7  & 47.1    
    \\
    & Adam & 65.2 & 28.9 & 36.9
    \\
    \midrule
    \multirow{5}{*}{SAM}
    & $\rho=0.01$ & 67.9 & 38.9 & 48.6\\
    & $\rho=0.02$ & 67.7 & 40.2 & 50.1\\
    & $\rho=0.03$ & {68.3} & 38.8 & 49.9\\
    & $\rho=0.04$ & 67.5 & 39.1 & 49.1\\
    & $\rho=0.05$ & \textbf{68.6} & \textbf{40.4} & \textbf{50.6}\\
    \midrule
    \multirow{2}{*}{$\ell_\infty$-AT} & $\epsilon=1/255$ & \red{50.8} & 50.2 & 50.4\\
    & $\epsilon=2/255$ & 49.2 & 48.7 & 49.0\\
    \midrule
    \multirow{3}{*}{$\ell_2$-AT} & $\epsilon=64/255$ & \red{60.3} & \textbf{56.9} & \textbf{58.4}\\
    & $\epsilon=128/255$ & 56.5 & 54.3 & 55.2\\
    & $\epsilon=255/255$ & 53.7 & 52.3 & 52.8\\
     \bottomrule
    \end{tabular}
    \vspace{-10pt}
    \label{tabel-seg-voc}
\end{table}

\textbf{Experimental results.}
The experiment results are summarized in Table~\ref{tabel-seg} and Table~\ref{tabel-seg-voc}, where we use a 10-step PGD attack under $\ell_\infty$-$\epsilon=1/255$ and $\ell_2$-$\epsilon=1$ for robustness evaluation, respectively. As shown in the table, models trained using SAM consistently achieve better segmentation performance than ST and significantly better than AT, and exhibit notably better robustness than ST, showcasing the robust generalization ability of SAM in various visual tasks. We also observed that the mIoU decrease of segmentation tasks is not as significant as the accuracy decrease of classification under adversarial attacks, which we assume is because the segmentation task is a highly interpretive task and many low-level features also play an important role in the segmentation results~\cite{ren2003learning,zhu2021learning}, making segmentation models inherently more robust than classification models.

\subsection{Text Classification}

Language models are also shown to be vulnerable against adversarial attacks~\cite{morris2020textattack,wei2024weighted}. While adversarial attack~\citep{zou2023universal,dong2023robust,wei2023jailbreak,zhang2024boosting} and defense~\citep{xie2023defending,piet2023jatmo,mo2024studious,wang2024theoretical} on Large Language Models (LLMs) have become emergent research topics recently, it is worth noting that adversarial training on LLMs may not be helpful and practical~\cite{jain2023baseline}, further underscoring the potential of deploying SAM a defense for language models.

\textbf{Experimental settings.}
In this part, we explore the effectiveness of SAM in enhancing the robustness of text sequence classification models against adversarial attacks. Utilizing the Rotten Tomatoes dataset~\cite{rotten} for sentiment analysis, we study the word-level adversarial robustness of models trained with different methods. Specifically, we consider two popular attacks: the Improved Genetic Algorithm (IGA)~\cite{iga} and the Particle Swarm Optimization (PSO)~\cite{pso}, both of which are officially supported by the TextAttack framework~\cite{morris2020textattack}. For the model under test, we selected the widely recognized \texttt{distilbert-base-uncased} model from Hugging Face's transformers library~\cite{DistilBERT}, fine-tuning it for 3 epochs on the target dataset. The optimization for ST was performed using the AdamW optimizer with a learning rate of \texttt{5e-5}, and the SAM optimization strategy was applied with the same base learning rate for AdamW and $\rho=0.05$. We also compare SAM with AT which incorporates adversarial examples identified in various adversarial attacks into the training dataset. Adversarial attacks such as IGA, PSO, and Textfooler~\cite{jin2020bert} have been chosen for data augmentation purposes. The official implementation in TextAttack provides specific details for training in this regard.

\textbf{Experimental results.}
The experimental results, summarized in~\cref{tab:text}, show the nuanced performance differences between SAM and ST. Although both SAM and ST achieved comparable natural accuracy (84.6$\%$ \textit{v.s.} 84.4$\%$), the distinction in attack success rates (ASR) is pronounced, \textit{e.g.} a reduced ASR of 17.1$\%$ under SAM compared to 22.7$\%$ under ST, indicating a significant improvement in adversarial robustness. As for AT, adding adversarial examples to the training set has been found to lower natural accuracy without significantly improving adversarial robustness. On average, the ASR for models trained with AT is 2.9\% higher with PSO attack and 2.6\% higher with IGA attack, indicating that adversarially trained models hurt overall performance against adversarial attacks, although they may be effective at defending specific types of attacks.
\begin{table}[h]
    \centering
    \vspace{-10pt}
    \caption{Accuracy and ASR comparison between SAM and ST for text sequence classification.}
    
    \begin{tabular}{c|c|cc}
    \toprule
    \multirow{2}{*}{Method} & Natural & PSO & IGA \\ & Accuracy ($\uparrow$) & ASR ($\downarrow$) & ASR ($\downarrow$) \\
    \midrule
    ST (AdamW) & 84.4 & 22.7 & 84.1
    \\
    \midrule
       SAM & \textbf{84.6} & \textbf{17.1} & \textbf{74.5}
    \\  
    \midrule
    AT (w/IGA) & 84.0	& 22.1 &	71.7
    \\
    AT (w/PSO) & 83.8	& 16.9	& 86.8
    \\
    AT (w/Textfooler) & 83.6	&21.1	&72.9
    \\
     \bottomrule
    \end{tabular}
    \vspace{-10pt}
    \label{tab:text}
\end{table}

These findings suggest that incorporating SAM can notably improve the adversarial robustness of text classification models without compromising natural performance, offering a promising avenue for future research and application to secure AI models against adversarial threats.

\section{Conclusion}
\label{Section 6}
In this paper, we reveal the duality relationship between Sharpness-Aware Minimization (SAM) and Adversarial Training (AT) and show that using SAM alone can improve adversarial robustness. 
We first intuitively illustrate that both SAM and AT can learn robust features by demonstrating the duality between weight and sample perturbations, and provide theoretical justifications to support these insights. We further conduct extensive experiments under various settings to show that SAM can improve robustness without sacrificing natural performance, while AT inevitably hurts natural generalization. These results uncover the scalability and practicality of using SAM to improve robustness without compromising accuracy on clean data. Based on this, we propose that SAM can be used as a lightweight alternative to AT when accuracy is a priority and improved robustness is preferred.

\section*{Impact Statement}
This work sheds light on how Sharpness-Aware Minimization (SAM), previously known for its clean accuracy improvements, can unexpectedly enhance the security of deep learning models by boosting their adversarial robustness. This finding presents a potential alternative to Adversarial Training (AT), a widely used but accuracy-sacrificing defense. By offering improved security without compromising clean accuracy, our work could significantly impact the deployment of reliable machine learning models in sensitive domains like healthcare, finance, and autonomous systems. However, responsible development and deployment practices are crucial, as robust models can also be misused. We emphasize the need for continued research into diverse defense mechanisms, fairness, interpretability, and theoretical underpinnings to ensure a secure and ethical future for machine learning.

\section*{Acknowledgement}

This work was sponsored by the Beijing Natural Science Foundation's Undergraduate Initiating Research Program (Grant No. QY23041) and the National Natural Science Foundation of China (Grant No. 62172019, 623B2001, 82371112, 62394311).

{
\bibliography{ref}

\begin{thebibliography}{83}
\providecommand{\natexlab}[1]{#1}
\providecommand{\url}[1]{\texttt{#1}}
\expandafter\ifx\csname urlstyle\endcsname\relax
  \providecommand{\doi}[1]{doi: #1}\else
  \providecommand{\doi}{doi: \begingroup \urlstyle{rm}\Url}\fi

\bibitem[Andriushchenko and Flammarion(2022)]{andriushchenko2022towards}
Maksym Andriushchenko and Nicolas Flammarion.
\newblock Towards understanding sharpness-aware minimization.
\newblock In \emph{ICML}, 2022.

\bibitem[Athalye et~al.(2018)Athalye, Carlini, and
  Wagner]{athalye2018obfuscated}
Anish Athalye, Nicholas Carlini, and David Wagner.
\newblock Obfuscated gradients give a false sense of security: Circumventing
  defenses to adversarial examples.
\newblock In \emph{ICML}, 2018.

\bibitem[Bahri et~al.(2021)Bahri, Mobahi, and Tay]{bahri2021sharpness}
Dara Bahri, Hossein Mobahi, and Yi~Tay.
\newblock Sharpness-aware minimization improves language model generalization.
\newblock \emph{arXiv preprint arXiv:2110.08529}, 2021.

\bibitem[Bai et~al.(2019)Bai, Feng, Wang, Dai, Xia, and Jiang]{bai2019hilbert}
Yang Bai, Yan Feng, Yisen Wang, Tao Dai, Shu-Tao Xia, and Yong Jiang.
\newblock Hilbert-based generative defense for adversarial examples.
\newblock In \emph{ICCV}, 2019.

\bibitem[Chaudhari et~al.(2019)Chaudhari, Choromanska, Soatto, LeCun, Baldassi,
  Borgs, Chayes, Sagun, and Zecchina]{chaudhari2019entropy}
Pratik Chaudhari, Anna Choromanska, Stefano Soatto, Yann LeCun, Carlo Baldassi,
  Christian Borgs, Jennifer Chayes, Levent Sagun, and Riccardo Zecchina.
\newblock Entropy-sgd: Biasing gradient descent into wide valleys.
\newblock \emph{Journal of Statistical Mechanics: Theory and Experiment}, 2019.

\bibitem[Chen et~al.(2022)Chen, Shao, Wang, Shang, Chen, Ji, and
  Wu]{chen2022bootstrap}
Huanran Chen, Shitong Shao, Ziyi Wang, Zirui Shang, Jin Chen, Xiaofeng Ji, and
  Xinxiao Wu.
\newblock Bootstrap generalization ability from loss landscape perspective.
\newblock In \emph{European Conference on Computer Vision}, pages 500--517.
  Springer, 2022.

\bibitem[Chen et~al.(2023{\natexlab{a}})Chen, Dong, Wang, Yang, Duan, Su, and
  Zhu]{chen2023robust}
Huanran Chen, Yinpeng Dong, Zhengyi Wang, Xiao Yang, Chengqi Duan, Hang Su, and
  Jun Zhu.
\newblock Robust classification via a single diffusion model.
\newblock \emph{arXiv preprint arXiv:2305.15241}, 2023{\natexlab{a}}.

\bibitem[Chen et~al.(2023{\natexlab{b}})Chen, Zhang, Dong, Yang, Su, and
  Zhu]{chen2023rethinking}
Huanran Chen, Yichi Zhang, Yinpeng Dong, Xiao Yang, Hang Su, and Jun Zhu.
\newblock Rethinking model ensemble in transfer-based adversarial attacks.
\newblock In \emph{The Twelfth International Conference on Learning
  Representations}, 2023{\natexlab{b}}.

\bibitem[Chen et~al.(2024)Chen, Dong, Shao, Hao, Yang, Su, and
  Zhu]{chen2024your}
Huanran Chen, Yinpeng Dong, Shitong Shao, Zhongkai Hao, Xiao Yang, Hang Su, and
  Jun Zhu.
\newblock Your diffusion model is secretly a certifiably robust classifier.
\newblock \emph{arXiv preprint arXiv:2402.02316}, 2024.

\bibitem[Chen et~al.(2017)Chen, Papandreou, Schroff, and
  Adam]{chen2017rethinking}
Liang-Chieh Chen, George Papandreou, Florian Schroff, and Hartwig Adam.
\newblock Rethinking atrous convolution for semantic image segmentation.
\newblock \emph{arXiv preprint arXiv:1706.05587}, 2017.

\bibitem[Chrabaszcz et~al.(2017)Chrabaszcz, Loshchilov, and
  Hutter]{chrabaszcz2017downsampled}
Patryk Chrabaszcz, Ilya Loshchilov, and Frank Hutter.
\newblock A downsampled variant of imagenet as an alternative to the cifar
  datasets.
\newblock \emph{arXiv preprint arXiv:1707.08819}, 2017.

\bibitem[Cohen et~al.(2019)Cohen, Rosenfeld, and Kolter]{cohen2019certified}
Jeremy~M Cohen, Elan Rosenfeld, and J.~Zico Kolter.
\newblock Certified adversarial robustness via randomized smoothing, 2019.

\bibitem[Croce and Hein(2020{\natexlab{a}})]{croce2020minimally}
F.~Croce and M.~Hein.
\newblock Minimally distorted adversarial examples with a fast adaptive
  boundary attack.
\newblock In \emph{ICML}, 2020{\natexlab{a}}.

\bibitem[Croce and Hein(2020{\natexlab{b}})]{croce2020reliable}
Francesco Croce and Matthias Hein.
\newblock Reliable evaluation of adversarial robustness with an ensemble of
  diverse parameter-free attacks.
\newblock In \emph{ICML}, 2020{\natexlab{b}}.

\bibitem[Croce and Hein(2021)]{croce2021mind}
Francesco Croce and Matthias Hein.
\newblock Mind the box: $l_1$-apgd for sparse adversarial attacks on image
  classifiers.
\newblock In \emph{ICML}, 2021.

\bibitem[Croce et~al.(2021)Croce, Andriushchenko, Sehwag, Debenedetti,
  Flammarion, Chiang, Mittal, and Hein]{croce2021robustbench}
Francesco Croce, Maksym Andriushchenko, Vikash Sehwag, Edoardo Debenedetti,
  Nicolas Flammarion, Mung Chiang, Prateek Mittal, and Matthias Hein.
\newblock Robustbench: a standardized adversarial robustness benchmark.
\newblock In \emph{NeurIPS}, 2021.

\bibitem[Dong et~al.(2023)Dong, Chen, Chen, Fang, Yang, Zhang, Tian, Su, and
  Zhu]{dong2023robust}
Yinpeng Dong, Huanran Chen, Jiawei Chen, Zhengwei Fang, Xiao Yang, Yichi Zhang,
  Yu~Tian, Hang Su, and Jun Zhu.
\newblock How robust is google's bard to adversarial image attacks?
\newblock In \emph{R0-FoMo: Robustness of Few-shot and Zero-shot Learning in
  Large Foundation Models}, 2023.

\bibitem[Du et~al.(2021)Du, Yan, Feng, Zhou, Zhen, Goh, and
  Tan]{du2021efficient}
Jiawei Du, Hanshu Yan, Jiashi Feng, Joey~Tianyi Zhou, Liangli Zhen, Rick
  Siow~Mong Goh, and Vincent~YF Tan.
\newblock Efficient sharpness-aware minimization for improved training of
  neural networks.
\newblock \emph{arXiv preprint arXiv:2110.03141}, 2021.

\bibitem[Du et~al.(2022)Du, Yan, Feng, Zhou, Zhen, Goh, and
  Tan]{du2022efficient}
Jiawei Du, Hanshu Yan, Jiashi Feng, Joey~Tianyi Zhou, Liangli Zhen, Rick
  Siow~Mong Goh, and Vincent Y.~F. Tan.
\newblock Efficient sharpness-aware minimization for improved training of
  neural networks.
\newblock \emph{arXiv preprint arXiv:2110.03141}, 2022.

\bibitem[Dziugaite and Roy(2017)]{dziugaite2017computing}
Gintare~Karolina Dziugaite and Daniel~M Roy.
\newblock Computing nonvacuous generalization bounds for deep (stochastic)
  neural networks with many more parameters than training data.
\newblock \emph{arXiv preprint arXiv:1703.11008}, 2017.

\bibitem[Everingham et~al.()Everingham, Van~Gool, Williams, Winn, and
  Zisserman]{voc-2012}
M.~Everingham, L.~Van~Gool, C.~K.~I. Williams, J.~Winn, and A.~Zisserman.
\newblock The {PASCAL} {V}isual {O}bject {C}lasses {C}hallenge 2012 {(VOC2012)}
  {R}esults.

\bibitem[Everingham et~al.(2015)Everingham, Eslami, Gool, Williams, Winn, and
  Zisserman]{mIoU}
Mark Everingham, S.~M.~Ali Eslami, Luc~Van Gool, Christopher K.~I. Williams,
  John~M. Winn, and Andrew Zisserman.
\newblock The pascal visual object classes challenge: {A} retrospective.
\newblock \emph{Int. J. Comput. Vis.}, 2015.

\bibitem[Foret et~al.(2020)Foret, Kleiner, Mobahi, and
  Neyshabur]{foret2020sharpness}
Pierre Foret, Ariel Kleiner, Hossein Mobahi, and Behnam Neyshabur.
\newblock Sharpness-aware minimization for efficiently improving
  generalization.
\newblock \emph{arXiv preprint arXiv:2010.01412}, 2020.

\bibitem[Goodfellow et~al.(2014)Goodfellow, Shlens, and
  Szegedy]{goodfellow2014explaining}
Ian~J Goodfellow, Jonathon Shlens, and Christian Szegedy.
\newblock Explaining and harnessing adversarial examples.
\newblock \emph{arXiv preprint arXiv:1412.6572}, 2014.

\bibitem[Goodfellow et~al.(2015)Goodfellow, Shlens, and
  Szegedy]{goodfellow2015explaining}
Ian~J. Goodfellow, Jonathon Shlens, and Christian Szegedy.
\newblock Explaining and harnessing adversarial examples.
\newblock \emph{arXiv preprint arXiv:1412.6572}, 2015.

\bibitem[Gould et~al.()Gould, Fulton, and Koller]{background_dataset}
Stephen Gould, Richard Fulton, and Daphne Koller.
\newblock Decomposing a scene into geometric and semantically consistent
  regions.
\newblock In \emph{ICCV 2009}.

\bibitem[He et~al.(2016{\natexlab{a}})He, Zhang, Ren, and Sun]{he2016deep}
Kaiming He, Xiangyu Zhang, Shaoqing Ren, and Jian Sun.
\newblock Deep residual learning for image recognition.
\newblock In \emph{CVPR}, 2016{\natexlab{a}}.

\bibitem[He et~al.(2016{\natexlab{b}})He, Zhang, Ren, and Sun]{he2016identity}
Kaiming He, Xiangyu Zhang, Shaoqing Ren, and Jian Sun.
\newblock Identity mappings in deep residual networks.
\newblock In \emph{ECCV}, 2016{\natexlab{b}}.

\bibitem[Hendrycks and Dietterich(2019)]{hendrycks2019benchmarking}
Dan Hendrycks and Thomas Dietterich.
\newblock Benchmarking neural network robustness to common corruptions and
  perturbations, 2019.

\bibitem[Hochreiter and Schmidhuber(1994)]{hochreiter1994simplifying}
Sepp Hochreiter and J{\"u}rgen Schmidhuber.
\newblock Simplifying neural nets by discovering flat minima.
\newblock \emph{NeurIPS}, 1994.

\bibitem[Hochreiter and Schmidhuber(1997)]{hochreiter1997flat}
Sepp Hochreiter and J{\"u}rgen Schmidhuber.
\newblock Flat minima.
\newblock \emph{Neural computation}, 1997.

\bibitem[Ilyas et~al.(2019)Ilyas, Santurkar, Tsipras, Engstrom, Tran, and
  Madry]{Ilyas2019AdversarialEA}
Andrew Ilyas, Shibani Santurkar, Dimitris Tsipras, Logan Engstrom, Brandon
  Tran, and Aleksander Madry.
\newblock Adversarial examples are not bugs, they are features.
\newblock In \emph{NeurIPS}, 2019.

\bibitem[Izmailov et~al.(2018)Izmailov, Podoprikhin, Garipov, Vetrov, and
  Wilson]{izmailov2018averaging}
Pavel Izmailov, Dmitrii Podoprikhin, Timur Garipov, Dmitry Vetrov, and
  Andrew~Gordon Wilson.
\newblock Averaging weights leads to wider optima and better generalization.
\newblock \emph{arXiv preprint arXiv:1803.05407}, 2018.

\bibitem[Jain et~al.(2023)Jain, Schwarzschild, Wen, Somepalli, Kirchenbauer,
  yeh Chiang, Goldblum, Saha, Geiping, and Goldstein]{jain2023baseline}
Neel Jain, Avi Schwarzschild, Yuxin Wen, Gowthami Somepalli, John Kirchenbauer,
  Ping yeh Chiang, Micah Goldblum, Aniruddha Saha, Jonas Geiping, and Tom
  Goldstein.
\newblock Baseline defenses for adversarial attacks against aligned language
  models, 2023.

\bibitem[Jin et~al.(2020)Jin, Jin, Zhou, and Szolovits]{jin2020bert}
Di~Jin, Zhijing Jin, Joey~Tianyi Zhou, and Peter Szolovits.
\newblock Is bert really robust? a strong baseline for natural language attack
  on text classification and entailment, 2020.

\bibitem[Keskar et~al.(2016)Keskar, Mudigere, Nocedal, Smelyanskiy, and
  Tang]{keskar2016large}
Nitish~Shirish Keskar, Dheevatsa Mudigere, Jorge Nocedal, Mikhail Smelyanskiy,
  and Ping Tak~Peter Tang.
\newblock On large-batch training for deep learning: Generalization gap and
  sharp minima.
\newblock \emph{arXiv preprint arXiv:1609.04836}, 2016.

\bibitem[Kim(2020)]{kim2020torchattacks}
Hoki Kim.
\newblock Torchattacks: A pytorch repository for adversarial attacks.
\newblock \emph{arXiv preprint arXiv:2010.01950}, 2020.

\bibitem[Kim et~al.(2022)Kim, Li, Hu, and Hospedales]{kim2022fisher}
Minyoung Kim, Da~Li, Shell~X Hu, and Timothy Hospedales.
\newblock Fisher sam: Information geometry and sharpness aware minimisation.
\newblock In \emph{ICML}, 2022.

\bibitem[Krizhevsky et~al.(2009)Krizhevsky, Hinton,
  et~al.]{krizhevsky2009learning}
Alex Krizhevsky, Geoffrey Hinton, et~al.
\newblock Learning multiple layers of features from tiny images.
\newblock 2009.

\bibitem[Kwon et~al.(2021)Kwon, Kim, Park, and Choi]{kwon2021asam}
Jungmin Kwon, Jeongseop Kim, Hyunseo Park, and In~Kwon Choi.
\newblock Asam: Adaptive sharpness-aware minimization for scale-invariant
  learning of deep neural networks.
\newblock In \emph{ICML}, 2021.

\bibitem[Liu et~al.(2022)Liu, Mai, Chen, Hsieh, and You]{liu2022towards}
Yong Liu, Siqi Mai, Xiangning Chen, Cho-Jui Hsieh, and Yang You.
\newblock Towards efficient and scalable sharpness-aware minimization.
\newblock In \emph{CVPR}, 2022.

\bibitem[Ma et~al.(2023)Ma, Pan, and massoud Farahmand]{ma2023understanding}
Avery Ma, Yangchen Pan, and Amir massoud Farahmand.
\newblock Understanding the robustness difference between stochastic gradient
  descent and adaptive gradient methods.
\newblock \emph{arXiv preprint arXiv:2308.06703}, 2023.

\bibitem[Madry et~al.(2017)Madry, Makelov, Schmidt, Tsipras, and
  Vladu]{madry2017towards}
Aleksander Madry, Aleksandar Makelov, Ludwig Schmidt, Dimitris Tsipras, and
  Adrian Vladu.
\newblock Towards deep learning models resistant to adversarial attacks.
\newblock \emph{arXiv preprint arXiv:1706.06083}, 2017.

\bibitem[Mi et~al.(2022)Mi, Shen, Ren, Zhou, Sun, Ji, and Tao]{mi2022make}
Peng Mi, Li~Shen, Tianhe Ren, Yiyi Zhou, Xiaoshuai Sun, Rongrong Ji, and
  Dacheng Tao.
\newblock Make sharpness-aware minimization stronger: A sparsified perturbation
  approach.
\newblock \emph{arXiv preprint arXiv:2210.05177}, 2022.

\bibitem[Mo et~al.(2024)Mo, Wang, Wei, and Wang]{mo2024studious}
Yichuan Mo, Yuji Wang, Zeming Wei, and Yisen Wang.
\newblock Studious bob fight back against jailbreaking via prompt adversarial
  tuning.
\newblock \emph{arXiv preprint arXiv:2402.06255}, 2024.

\bibitem[Morris et~al.(2020)Morris, Lifland, Yoo, Grigsby, Jin, and
  Qi]{morris2020textattack}
John Morris, Eli Lifland, Jin~Yong Yoo, Jake Grigsby, Di~Jin, and Yanjun Qi.
\newblock Textattack: A framework for adversarial attacks, data augmentation,
  and adversarial training in nlp.
\newblock In \emph{EMNLP}, 2020.

\bibitem[Mueller et~al.(2023)Mueller, Vlaar, Rolnick, and
  Hein]{mueller2023normalization}
Maximilian Mueller, Tiffany Vlaar, David Rolnick, and Matthias Hein.
\newblock Normalization layers are all that sharpness-aware minimization needs.
\newblock \emph{arXiv preprint arXiv:2306.04226}, 2023.

\bibitem[Neyshabur et~al.(2017)Neyshabur, Bhojanapalli, McAllester, and
  Srebro]{neyshabur2017exploring}
Behnam Neyshabur, Srinadh Bhojanapalli, David McAllester, and Nati Srebro.
\newblock Exploring generalization in deep learning.
\newblock \emph{NeurIPS}, 30, 2017.

\bibitem[Pang and Lee(2005)]{rotten}
Bo~Pang and Lillian Lee.
\newblock Seeing stars: Exploiting class relationships for sentiment
  categorization with respect to rating scales.
\newblock In \emph{ACL}, 2005.

\bibitem[Pang et~al.(2020)Pang, Yang, Dong, Su, and Zhu]{pang2020bag}
Tianyu Pang, Xiao Yang, Yinpeng Dong, Hang Su, and Jun Zhu.
\newblock Bag of tricks for adversarial training.
\newblock \emph{arXiv preprint arXiv:2010.00467}, 2020.

\bibitem[Papernot et~al.(2016)Papernot, McDaniel, Wu, Jha, and
  Swami]{papernot2016distillation}
Nicolas Papernot, Patrick McDaniel, Xi~Wu, Somesh Jha, and Ananthram Swami.
\newblock Distillation as a defense to adversarial perturbations against deep
  neural networks.
\newblock In \emph{SP}, 2016.

\bibitem[Piet et~al.(2023)Piet, Alrashed, Sitawarin, Chen, Wei, Sun, Alomair,
  and Wagner]{piet2023jatmo}
Julien Piet, Maha Alrashed, Chawin Sitawarin, Sizhe Chen, Zeming Wei, Elizabeth
  Sun, Basel Alomair, and David Wagner.
\newblock Jatmo: Prompt injection defense by task-specific finetuning.
\newblock \emph{arXiv preprint arXiv:2312.17673}, 2023.

\bibitem[Pomponi et~al.(2022)Pomponi, Scardapane, and Uncini]{Pomponi_2022}
Jary Pomponi, Simone Scardapane, and Aurelio Uncini.
\newblock Pixle: a fast and effective black-box attack based on rearranging
  pixels.
\newblock In \emph{IJCNN}, 2022.

\bibitem[Ren and Malik(2003)]{ren2003learning}
Ren and Malik.
\newblock Learning a classification model for segmentation.
\newblock In \emph{ICCV}, 2003.

\bibitem[Rice et~al.(2020)Rice, Wong, and Kolter]{rice2020overfitting}
Leslie Rice, Eric Wong, and J.~Zico Kolter.
\newblock Overfitting in adversarially robust deep learning, 2020.

\bibitem[Sandler et~al.(2018)Sandler, Howard, Zhu, Zhmoginov, and
  Chen]{mobilenetv2}
Mark Sandler, Andrew~G. Howard, Menglong Zhu, Andrey Zhmoginov, and
  Liang{-}Chieh Chen.
\newblock Mobilenetv2: Inverted residuals and linear bottlenecks.
\newblock In \emph{{CVPR}}, 2018.

\bibitem[Sanh et~al.(2019)Sanh, Debut, Chaumond, and Wolf]{DistilBERT}
Victor Sanh, Lysandre Debut, Julien Chaumond, and Thomas Wolf.
\newblock Distilbert, a distilled version of bert: smaller, faster, cheaper and
  lighter.
\newblock \emph{ArXiv}, abs/1910.01108, 2019.

\bibitem[Shafahi et~al.(2019)Shafahi, Najibi, Ghiasi, Xu, Dickerson, Studer,
  Davis, Taylor, and Goldstein]{shafahi2019adversarial}
Ali Shafahi, Mahyar Najibi, Mohammad~Amin Ghiasi, Zheng Xu, John Dickerson,
  Christoph Studer, Larry~S Davis, Gavin Taylor, and Tom Goldstein.
\newblock Adversarial training for free!
\newblock \emph{NeurIPS}, 2019.

\bibitem[Szegedy et~al.(2013)Szegedy, Zaremba, Sutskever, Bruna, Erhan,
  Goodfellow, and Fergus]{szegedy2013intriguing}
Christian Szegedy, Wojciech Zaremba, Ilya Sutskever, Joan Bruna, Dumitru Erhan,
  Ian Goodfellow, and Rob Fergus.
\newblock Intriguing properties of neural networks.
\newblock \emph{arXiv preprint arXiv:1312.6199}, 2013.

\bibitem[Touvron et~al.(2021)Touvron, Cord, Douze, Massa, Sablayrolles, and
  Jégou]{touvron2021training}
Hugo Touvron, Matthieu Cord, Matthijs Douze, Francisco Massa, Alexandre
  Sablayrolles, and Hervé Jégou.
\newblock Training data-efficient image transformers distillation through
  attention.
\newblock \emph{arXiv preprint arXiv:2012.12877}, 2021.

\bibitem[Tsipras et~al.(2018)Tsipras, Santurkar, Engstrom, Turner, and
  Madry]{tsipras2018robustness}
Dimitris Tsipras, Shibani Santurkar, Logan Engstrom, Alexander Turner, and
  Aleksander Madry.
\newblock Robustness may be at odds with accuracy.
\newblock \emph{arXiv preprint arXiv:1805.12152}, 2018.

\bibitem[Wang et~al.(2021)Wang, Jin, Yang, and He]{iga}
Xiaosen Wang, Hao Jin, Yichen Yang, and Kun He.
\newblock Natural language adversarial defense through synonym encoding.
\newblock \emph{arXiv preprint arXiv:1909.06723}, 2021.

\bibitem[Wang et~al.(2023)Wang, Li, Yang, Lin, and Wang]{wang2024balance}
Yifei Wang, Liangchen Li, Jiansheng Yang, Zhouchen Lin, and Yisen Wang.
\newblock Balance, imbalance, and rebalance: Understanding robust overfitting
  from a minimax game perspective.
\newblock \emph{NeurIPS}, 2023.

\bibitem[Wang et~al.(2024)Wang, Wu, Wei, Jegelka, and
  Wang]{wang2024theoretical}
Yifei Wang, Yuyang Wu, Zeming Wei, Stefanie Jegelka, and Yisen Wang.
\newblock A theoretical understanding of self-correction through in-context
  alignment.
\newblock \emph{arXiv preprint arXiv:2405.18634}, 2024.

\bibitem[Wei et~al.(2023{\natexlab{a}})Wei, Guo, and
  Wang]{wei2023characterizing}
Zeming Wei, Yiwen Guo, and Yisen Wang.
\newblock Characterizing robust overfitting in adversarial training via
  cross-class features.
\newblock \emph{OpenReview preprint}, 2023{\natexlab{a}}.

\bibitem[Wei et~al.(2023{\natexlab{b}})Wei, Wang, Guo, and Wang]{wei2023cfa}
Zeming Wei, Yifei Wang, Yiwen Guo, and Yisen Wang.
\newblock Cfa: Class-wise calibrated fair adversarial training.
\newblock In \emph{CVPR}, 2023{\natexlab{b}}.

\bibitem[Wei et~al.(2023{\natexlab{c}})Wei, Wang, and Wang]{wei2023jailbreak}
Zeming Wei, Yifei Wang, and Yisen Wang.
\newblock Jailbreak and guard aligned language models with only few in-context
  demonstrations.
\newblock \emph{arXiv preprint arXiv:2310.06387}, 2023{\natexlab{c}}.

\bibitem[Wei et~al.(2024)Wei, Zhang, Zhang, and Sun]{wei2024weighted}
Zeming Wei, Xiyue Zhang, Yihao Zhang, and Meng Sun.
\newblock Weighted automata extraction and explanation of recurrent neural
  networks for natural language tasks.
\newblock \emph{Journal of Logical and Algebraic Methods in Programming}, 2024.

\bibitem[Wong et~al.(2020)Wong, Rice, and Kolter]{wong2020fast}
Eric Wong, Leslie Rice, and J~Zico Kolter.
\newblock Fast is better than free: Revisiting adversarial training.
\newblock \emph{arXiv preprint arXiv:2001.03994}, 2020.

\bibitem[Wu et~al.(2020)Wu, Xia, and Wang]{wu2020adversarial}
Dongxian Wu, Shu-Tao Xia, and Yisen Wang.
\newblock Adversarial weight perturbation helps robust generalization.
\newblock In \emph{NeurIPS}, 2020.

\bibitem[Xiao et~al.(2018)Xiao, Zhu, Li, He, Liu, and Song]{xiao2018spatially}
Chaowei Xiao, Jun-Yan Zhu, Bo~Li, Warren He, Mingyan Liu, and Dawn Song.
\newblock Spatially transformed adversarial examples.
\newblock \emph{arXiv preprint arXiv:1801.02612}, 2018.

\bibitem[Xie et~al.(2019)Xie, Wu, Maaten, Yuille, and He]{xie2019feature}
Cihang Xie, Yuxin Wu, Laurens van~der Maaten, Alan~L Yuille, and Kaiming He.
\newblock Feature denoising for improving adversarial robustness.
\newblock In \emph{CVPR}, 2019.

\bibitem[Xie et~al.(2023)Xie, Yi, Shao, Curl, Lyu, Chen, Xie, and
  Wu]{xie2023defending}
Yueqi Xie, Jingwei Yi, Jiawei Shao, Justin Curl, Lingjuan Lyu, Qifeng Chen,
  Xing Xie, and Fangzhao Wu.
\newblock Defending chatgpt against jailbreak attack via self-reminders.
\newblock \emph{Nature Machine Intelligence}, 2023.

\bibitem[Xu et~al.(2021)Xu, Liu, Li, Jain, and Tang]{xu2021robust}
Han Xu, Xiaorui Liu, Yaxin Li, Anil Jain, and Jiliang Tang.
\newblock To be robust or to be fair: Towards fairness in adversarial training.
\newblock In \emph{ICML}, 2021.

\bibitem[Yu et~al.(2022{\natexlab{a}})Yu, Han, Gong, Shen, Ge, Du, and
  Liu]{yu2022robust}
Chaojian Yu, Bo~Han, Mingming Gong, Li~Shen, Shiming Ge, Bo~Du, and Tongliang
  Liu.
\newblock Robust weight perturbation for adversarial training.
\newblock \emph{arXiv preprint arXiv:2205.14826}, 2022{\natexlab{a}}.

\bibitem[Yu et~al.(2022{\natexlab{b}})Yu, Han, Gong, Shen, Ge, Du, and
  Liu]{yu2022robust2}
Chaojian Yu, Bo~Han, Mingming Gong, Li~Shen, Shiming Ge, Bo~Du, and Tongliang
  Liu.
\newblock Robust weight perturbation for adversarial training.
\newblock \emph{arXiv preprint arXiv:2205.14826}, 2022{\natexlab{b}}.

\bibitem[Zagoruyko and Komodakis(2016)]{DBLP:journals/corr/ZagoruykoK16}
Sergey Zagoruyko and Nikos Komodakis.
\newblock Wide residual networks.
\newblock \emph{arXiv preprint arXiv:1605.07146}, 2016.

\bibitem[Zang et~al.(2020)Zang, Qi, Yang, Liu, Zhang, Liu, and Sun]{pso}
Yuan Zang, Fanchao Qi, Chenghao Yang, Zhiyuan Liu, Meng Zhang, Qun Liu, and
  Maosong Sun.
\newblock Word-level textual adversarial attacking as combinatorial
  optimization.
\newblock In \emph{ACL}, 2020.

\bibitem[Zhang et~al.(2019)Zhang, Yu, Jiao, Xing, El~Ghaoui, and
  Jordan]{zhang2019theoretically}
Hongyang Zhang, Yaodong Yu, Jiantao Jiao, Eric Xing, Laurent El~Ghaoui, and
  Michael Jordan.
\newblock Theoretically principled trade-off between robustness and accuracy.
\newblock In \emph{ICML}, 2019.

\bibitem[Zhang and Wei(2024)]{zhang2024boosting}
Yihao Zhang and Zeming Wei.
\newblock Boosting jailbreak attack with momentum.
\newblock In \emph{ICLR 2024 Workshop on Reliable and Responsible Foundation
  Models}, 2024.

\bibitem[Zhu et~al.(2021)Zhu, Ji, Zhu, Gan, Wu, and Yan]{zhu2021learning}
Lanyun Zhu, Deyi Ji, Shiping Zhu, Weihao Gan, Wei Wu, and Junjie Yan.
\newblock Learning statistical texture for semantic segmentation.
\newblock In \emph{CVPR}, 2021.

\bibitem[Zhu et~al.(2023)Zhu, He, Chen, Song, and Tao]{zhu2023decentralized}
Tongtian Zhu, Fengxiang He, Kaixuan Chen, Mingli Song, and Dacheng Tao.
\newblock Decentralized sgd and average-direction sam are asymptotically
  equivalent.
\newblock \emph{arXiv preprint arXiv:2306.02913}, 2023.

\bibitem[Zou et~al.(2023)Zou, Wang, Kolter, and Fredrikson]{zou2023universal}
Andy Zou, Zifan Wang, J.~Zico Kolter, and Matt Fredrikson.
\newblock Universal and transferable adversarial attacks on aligned language
  models, 2023.

\end{thebibliography}
\bibliographystyle{plainnat}
}

\newpage
\appendix
\onecolumn
\section{Proofs}
\newcommand{\ud}{\text{d}}
\label{proofs}

In this section, we provide all the proof for the theorems. To start with, we introduce a property of our model in the data distribution, which has been proved and used in a series of previous works~\cite{tsipras2018robustness,Ilyas2019AdversarialEA,xu2021robust}:

\textbf{Lemma~\cite{tsipras2018robustness}} In the presented model and data distribution, given $w_1$, the optimal solution for the optimization objective will assign equal weight to all the non-robust features, \textit{i.e.} $w_2=w_3=\cdots=w_{n+1}$.

The conclusion is proved in Lemma D.1 in \citet{tsipras2018robustness}. Based on this lemma, since we only focus on the ratio $W_R = \frac{w_1}{w_2+w_3+\cdots+w_{n+1}}$, we can further assume $w_2=w_3=\cdots=w_{n+1}=1$ without loss of generalization.
\subsection{Proof for Theorem 4.1}
\begin{proof}
As $x_1$ has been chosen to be in $\pm 1$, the perturbation over $x_1$ has no influence on it, and we can just ignore it.
Therefore, to attack the classifier by a bias $\boldsymbol\delta$, to make the accuracy as small as possible, an intuitive idea is to set $\boldsymbol\delta$ which minimizes the expectation of $x_i(i = 2,\cdots,n+1),$ which made the standard accuracy smaller.
In fact, the expected accuracy is monotonically increasing about each $\delta_i (i = 2,\cdots,n+1).$ 
Thus, choosing $\delta = (0,-\epsilon , \cdots, -\epsilon )$ can be the best adversarial attack vector for any $w>0$.
In this situation, this equals $x_i'(i = 2,\cdots, n+1) \sim \mathcal{N}(\eta-\epsilon ,n).$ 

Thus, $R_A$ can be rewritten to be 
\begin{equation}
\begin{aligned}
R_A&=\mathbb{E}_{\mathbf{x},y\sim D}\mathbb{P}(w_1 x_1+\sum_{i=2}^{n+1}x_i-n\epsilon>0)\\
&=p\Phi((w_1+(\eta-\epsilon) n)/\sqrt{n}) + (1-p)\Phi((-w_1+(\eta-\epsilon) n)/\sqrt{n}),
\end{aligned}
\end{equation}
where $\Phi(\cdot)$ is the cumulative distribution function of a standard normal distribution.

Since $W_R$ is obviously a monotonic increasing function of $w_1$, we only need to prove that $RA$ in the above equation is a monotonic increasing function of $w_1$. To show this, we take a partial derivative of $w_1$ in $RA$:
\begin{equation}
\frac{\partial RA}{\partial w_1}=p\exp({-(w_1+(\eta-\epsilon)n)^2/2n})/\sqrt{2\pi n}-(1-p)\exp({-(w_1-(\eta-\epsilon)n)^2/2n})/\sqrt{2\pi n}.
\end{equation}
Simplifying this, we get
\begin{equation}
\frac{\partial R_A}{\partial w_1}=g(w_1)\cdot (1-\frac{1-p}{p}\exp(2w_1(\eta-\epsilon))),
\end{equation}
where $g(w_1)=\frac{p}{\sqrt{2\pi n}}\exp({-(w_1+(\eta-\epsilon)n)^2/2n})$ is positive . When $0<W_R<W_R^{AT}$, or equivalantly $0<w_1<\frac{\ln p - \ln (1-p)}{2(\eta-\epsilon)}$, we can easily know that $(1-\frac{1-p}{p}\exp(2w_1(\eta-\epsilon)))>0$, which implies that $\frac{\partial R_A}{\partial w_1}>0$, and further means that RA is a monotonic increasing function of $W_R$.
\end{proof}

\subsection{Proof for Theorem 4.2}
\begin{proof}
Due to symmetry, we only need to calculate the case of $y=1$ without loss of generality. From the distribution, we can easily derive that $x_2 + \cdots + x_{n+1} \sim \mathcal{N}(\eta n,n)$.

Thus, since $w_2=w_3=\cdots w_{n+1}=1$ are assumed to be fixed, we can know that the best parameter $w_1$ satisfies\begin{equation}\begin{aligned}
   w_1^*=&\underset{w_1}{\text{argmax}}\;\mathbb{E}_{\boldsymbol x.y \sim \mathcal{D}} \mathbf{1}_{f_{\boldsymbol w}(\boldsymbol x) = y}\\
=& \underset{w_1}{\text{argmax}}\; p\Pr(x_2 + \cdots + x_{n+1} > -w_1) + (1-p)\Pr(x_2 + \cdots + x_{n+1} > w_1)\\
=&  \underset{w_1}{\text{argmax}}\;\frac{p}{\sqrt{2\pi n}}\int_{-w_1}^\infty e^{-(t-\eta n)^2/2n}\ud t +\frac{1-p}{\sqrt{2\pi n}}\int_{w_1}^\infty e^{-(t-\eta n)^2/2n}\ud t\\
:=& \underset{w_1}{\text{argmax}}\;u(w_1).
\end{aligned}\label{eq7}\end{equation}

Then, the best parameter $w_1$ can be derived by $\ud u/\ud w_1 = 0.$ The derivative is \begin{equation}\frac{\ud u}{\ud w_1} = \frac{p}{\sqrt{2\pi n}} e^{-(w_1+\eta n)^2/2n} - \frac{1-p}{\sqrt{2\pi n}}e^{-(w_1-\eta n)^2/2n} = 0.\end{equation}
Solving this, we get the optimal value of $w_1$ is
\begin{equation}w_1^* = \frac{\ln p - \ln (1-p)}{2\eta}.\end{equation}
Therefore, $W_R^*$ under the optimal value of $\boldsymbol{w}$ is
\begin{equation}\label{Wst}
W_{R}^*=\frac{\ln p-\ln (1-p)}{2n\eta}.
\end{equation}
\end{proof}
\subsection{Proof for Theorem 4.3}

\begin{proof}As $x_1$ has been chosen to be in $\pm 1$, the perturbation over $x_1$ has no influence on it, and we can just ignore it.
Therefore, to attack the classifier by a bias $\boldsymbol\delta$, to make the accuracy as small as possible, an intuitive idea is to set $\boldsymbol\delta$ which minimizes the expectation of $x_i(i = 2,\cdots,n+1),$ which made the standard accuracy smaller.
In fact, the expected accuracy is monotonically increasing about each $\delta_i (i = 2,\cdots,n+1).$ 
Thus, choosing $\delta = (0,-\epsilon , \cdots, -\epsilon )$ can be the best adversarial attack vector for any $w>0$.
In this situation, this equals $x_i'(i = 2,\cdots, n+1) \sim \mathcal{N}(\eta-\epsilon ,n).$ 
Therefore, similar to equation (\ref{eq7}), we can derive the train accuracy which is 
\begin{equation}
    v(w) = p\Phi((w+(\eta-\epsilon) n)/\sqrt{n}) + (1-p)\Phi((-w+(\eta-\epsilon) n)/\sqrt{n}).
\end{equation}

Here $\Phi(\cdot)$ is the cumulative distribution function of a standard normal distribution. Now we only need to solve equation $dv/dw = 0$.
Through simple computation, this derives that
\begin{equation}\begin{aligned}    
   & p\exp({-(w_1+(\eta-\epsilon)n)^2/2n})/\sqrt{2\pi n}
   = (1-p)\exp({-(w_1-(\eta-\epsilon)n)^2/2n})/\sqrt{2\pi n}.
\end{aligned}
\end{equation} 
Solving this equation, we finally get the optimal value for $w_1$ to be \begin{equation}w_1^{AT} = \frac{\ln p - \ln (1-p)}{2(\eta-\epsilon)}.\end{equation}

Therefore, $W_R^{AT}$ under the optimal value of $w_1$ is
\begin{equation}\label{Wat}
W_R^{AT}=\frac{\ln p - \ln (1-p)}{2n(\eta-\epsilon)}.
\end{equation}
\end{proof}

\subsection{Proof for Theorem 4.4}

\textit{Proof.} 
Define the expected clean accuracy function 
\begin{equation}u(w) = \frac{p}{\sqrt{n}}\Phi((w+\eta n)/\sqrt{n}) + \frac{(1-p)}{\sqrt{n}}\Phi((-w+\eta n)/\sqrt{n}),\end{equation}
where $\Phi(\cdot)$ is the cumulative distribution function of a standard normal distribution and $w\in \mathbb{R}$. The derivative is
\begin{equation}\label{eq14}
    \begin{aligned}\frac{\ud u(w)}{\ud w}=&\frac{p}{\sqrt{2\pi n}}e^{-(w+\eta n)^{2}/2n}-\frac{1-p}{\sqrt{2\pi n}}e^{-(w-\eta n)^{2}/2n}\\
    =&\frac{1}{\sqrt{2\pi n}}e^{-(w^{2}+\eta^{2}n^{2})/n}\big{(}pe^{-w\eta}-(1-p)e^{w\eta}\big{)}.
    \end{aligned}
\end{equation}
From the theorem \ref{theorem ST} for standard training, we know $u(w)$ has only one global minimum $w_1^*= \frac{\ln p - \ln (1-p)}{2\eta}$ for which $\frac{\ud u(w)}{\ud w}|_{w=w_1^*}=0$. Thus, from (\ref{eq14}) we know that $\frac{d u(w)}{w}<0$ if $w>w_1^*$ and $\frac{d u(w)}{w}>0$ if $w<w_1^*$.

In the SAM algorithm where we set $\lambda=0$ and with $\epsilon$ given, we know that 
\begin{equation}
    w_1^{SAM} = \underset{w}{\text{argmax}}\min_{\delta \in [-\epsilon,\epsilon] }u(w+\delta).
\end{equation}

It is easy for us to know that state that 
\begin{equation}
    \min_{\delta \in [-\epsilon,\epsilon] }u(w_1^{SAM}+\delta)=\min \big{\{}u(w_1^{SAM}-\epsilon) , u(w_1^{SAM}+\epsilon)\big{\}}.
\end{equation}

If $u(w_1^{SAM}-\epsilon) > u(w_1^{SAM}+\epsilon),$ then $w_1^{SAM}+\epsilon>w_1^*$. Since $\frac{\ud u(w)}{\ud w}$ is continuous and locally bounded, there exists $\delta_{0}>0$ such that $u(w_1^{SAM}-\epsilon-\delta_0)>u(w_1^{SAM}+\epsilon)$ and $u(w_1^{SAM}+\epsilon-\delta_0)>u(w_1^{SAM}+\epsilon)$
 Thus, we have
\begin{equation}
\begin{aligned}
    &\min \big{\{}u(w_1^{SAM}-\epsilon-\delta_0) , u(w_1^{SAM}+\epsilon-\delta_0)\big{\}}>\min \big{\{}u(w_1^{SAM}-\epsilon) , u(w_1^{SAM}+\epsilon)\big{\}}.
    \end{aligned}
\end{equation}
Therefore
\begin{equation}
    \min_{\delta \in [-\epsilon,\epsilon] }u\big{(}(w_1^{SAM}-\delta_0)+\delta\big{)}>\min_{\delta \in [-\epsilon,\epsilon] }u(w_1^{SAM}+\delta),
\end{equation}
which means that $w_1^{SAM}$ is not the optimal value we want.
Similarly we can disprove that $u(w_1^{SAM}-\epsilon) <u(w_1^{SAM}+\epsilon).$ Thus, $u(w_1^{SAM}-\epsilon)=u(w_1^{SAM}+\epsilon).$

From this, we know that
\begin{equation}
    \int_{w_1^{SAM}-\epsilon}^{w_1^*}\frac{\ud u(w)}{\ud w}\ud w=-\int_{w_{1}^*}^{w_1^{SAM}+\epsilon}\frac{\ud u(w)}{\ud w}\ud w.
\end{equation}
Using (\ref{eq14}), we have
\begin{equation}\label{eq20}
\begin{aligned}
    &\int_{w_1^{SAM}-\epsilon}^{w_1^*}e^{-w^{2}}\big{(}pe^{-w\eta}-(1-p)e^{w\eta}\big{)}\ud w\\
    =&-\int_{w_{1}^*}^{w_1^{SAM}+\epsilon}e^{-w^{2}}\big{(}pe^{-w\eta}-(1-p)e^{w\eta}\big{)}\ud w.
\end{aligned}
\end{equation}
If $w_1^{SAM}\leq w_1^*$, define $h=w_1^{SAM}+\epsilon-w_1^*$. Thus, $h\leq\epsilon\leq-w_1^{SAM}+\epsilon+w_1^*$. In (\ref{eq20}), we have
\begin{equation}
\begin{aligned}\label{eq21}
    &\int_{-h}^0 e^{-(w_1^*+w)^{2}}\big{(}pe^{-(w_1^*+w)\eta}-(1-p)e^{(w_1^*+w)\eta}\big{)}\ud w\\
    \leq &\int_{w_1^{SAM}-\epsilon}^{w_1^*}e^{-w^{2}}\big{(}pe^{-w\eta}-(1-p)e^{w\eta}\big{)}\ud w\\
    =&-\int_{w_{1}^*}^{w_1^{SAM}+\epsilon}e^{-w^{2}}\big{(}pe^{-w\eta}-(1-p)e^{w\eta}\big{)}\ud w\\
    =&-\int_{0}^{h}e^{-(w_1^*+w)^{2}}\big{(}pe^{-(w_1^*+w)\eta}-(1-p)e^{(w_1^*+w)\eta}\big{)}\ud w\\
    =&\int_{-h}^{0}e^{-(w_1^*-w)^{2}}\big{(}-pe^{-(w_1^*-w)\eta}+(1-p)e^{(w_1^*-w)\eta}\big{)}\ud w.
\end{aligned}
\end{equation}
Since $w_1^*>0$, we can know that $e^{-(w_1^*+w)^{2}}>e^{-(w_1^*-w)^{2}}>0$ for $w\in[-h,0)$.

For the function $r(v):=\big{(}pe^{-v \eta}-(1-p)e^{v\eta}\big{)}$ is monotonically decreasing and has one zero point $w_1^*$, thus $pe^{-(w_1^*+w)\eta}-(1-p)e^{(w_1^*+w)\eta}>0$ and $-pe^{-(w_1^*-w)\eta}+(1-p)e^{(w_1^*-w)\eta}>0$ for $w\in [-h,0]$. And
\begin{equation}
\begin{aligned}\label{eq22}
    &pe^{-(w_1^*+w)\eta}-(1-p)e^{(w_1^*+w)\eta}-\big{(}-pe^{-(w_1^*-w)\eta}+(1-p)e^{(w_1^*-w)\eta}\big{)}\\
    =&\big{(}pe^{-w_1^* \eta}-(1-p)e^{w_1^*\eta}\big{)}(e^{w\eta}+e^{-w\eta})\\
    =&0.
\end{aligned}
\end{equation}
Therefore, $\forall w\in [-h,0)$,
\begin{equation}
    \begin{aligned}
        & e^{-(w_1^*+w)^{2}}\big{(}pe^{-(w_1^*+w)\eta}-(1-p)e^{(w_1^*+w)\eta}\big{)}
        > e^{-(w_1^*-w)^{2}}\big{(}-pe^{-(w_1^*-w)\eta}+(1-p)e^{(w_1^*-w)\eta}\big{)}.
    \end{aligned}
\end{equation}
Combining this with (\ref{eq21}), we reach a contradiction. 

Thus, $w_1^{SAM}>w_1^*$. Applying this to the definition of $W_R^{SAM}$ and $W_R^*$, we can easily obtain the result that 
\begin{equation}
W_R^{SAM}>W_R^*.
\end{equation}$\hfill \square$

\subsection{Proof for Theorem 4.5}

\begin{proof} 
We proceed our proof from (\ref{eq20}). Since we have proven that $w_1^{SAM}>w_1^*$, we suppose that $h=w_1^*-w_1^{SAM}+\epsilon<\epsilon$. Therefore, we can derive from (\ref{eq20}) and the definition of h that

\begin{equation}
    \begin{aligned}
        0=&\int_{-h}^0 e^{-(w_1^*+w)^2}(pe^{-(w_1^*+w)\eta}-(1-p)e^{(w_1^*+w)\eta})\ud w\\&-\int_{-h}^0 e^{-(w_1^*-w)^2}(-pe^{-(w_1^*-w)\eta}+(1-p)e^{(w_1^*-w)\eta})\ud w\\
        &-\int_{h}^{2\epsilon-h} e^{-(w_1^*+w)^2}(-pe^{-(w_1^*+w)\eta}+(1-p)e^{(w_1^*+w)\eta})\ud w.
    \end{aligned}
\end{equation}

Since we only focus on $h$, we consider omitting the $o(h)$ terms in the calculation. To be more specific, $o(w^2)$ term in the integral symbol $\int_{-h}^0$ can be omitted, and $o(w)$ or $o(h)$ term for $w$ and $h$ in the integral symbol $\int_{h}^{2\epsilon-2h}$ can also be omitted.\footnote{The validity of abandoning these high order terms can be seen from the result of the calculation, which shows that $\frac{h-\epsilon}{h}\to 0$.}

Combined with the proof in (\ref{eq22}) and the definition of $w_1^*$, and abandoning the high order terms, we can calculate the right-hand side as follows.
\begin{equation}\label{eq25}
    \begin{aligned}
        RHS=&\int_{-h}^0 \big{(}e^{-(w_1^*+w)^2}-e^{-(w_1^*-w)^2}\big{)}\cdot
        \big{(}pe^{-(w_1^*+w)\eta}-(1-p)e^{(w_1^*+w)\eta}\big{)}\ud w
        \\&+\int_{h}^{2\epsilon-h} e^{-(w_1^*+w)^2}\big{(}-pe^{-(w_1^*+w)\eta}+(1-p)e^{(w_1^*+w)\eta}\big{)}\ud w\\\approx
        &\int_{-h}^0 e^{-(w_1^*)^2}\big{(}1-2w_1^* w-1-2w_1^* w+o(w)\big{)}\cdot
        \big{(}pe^{-w_1^* \eta}(1-w\eta)-(1-p)e^{w_1^*\eta}(1+w\eta)\big{)}\ud w\\&-
        \int_{0}^{2\epsilon-2h} e^{-(w_1^*)^2}(1-2w_1^*h)\cdot
        \big{(}-pe^{-w_1^*\eta}(1-h\eta)+(1-p)e^{w_1^*\eta}(1+h\eta)\big{)}\ud w\\\approx
        &4e^{-(w_1^*)^2}w_1^*\int_{-h}^0 w^2\eta\big{(}-pe^{-w_1^* \eta}-(1-p)e^{w_1^*\eta}\big{)}\ud w-e^{-(w_1^*)^2}\int_{0}^{2\epsilon-2h}\eta h\big{(}pe^{-w_1^*\eta}+(1-p)e^{w_1^*\eta}\big{)}\ud w\\
        \approx& \frac{4}{3}e^{-(w_1^*)^2}w_1^*\big{(}pe^{-w_1^* \eta}+(1-p)e^{w_1^*\eta}\big{)}h^3
        -2e^{-(w_1^*)^2}(\epsilon-h)\eta h\big{(}pe^{-w_1^*\eta}+(1-p)e^{w_1^*\eta}\big{)}\\\approx
        & \frac{2}{3}e^{-(w_1^*)^2}\big{(}pe^{-w_1^*\eta}+(1-p)e^{w_1^*\eta}\big{)}\eta\big{(}2w_1^*h^2-3(\epsilon-h)\big{)}.
    \end{aligned}
\end{equation}
Since $RHS=0$, by solving the last equality in (\ref{eq25}) we get that
\begin{equation}
    \frac{\epsilon-h}{h}=\frac{2}{3}w_1^*h= o(1).
\end{equation}
Thus, the calculation and the abandoning of high-order terms in the calculation above are valid.
Since $h\to\epsilon$, we have
\begin{equation}
\begin{aligned}
    \epsilon-h=&\frac{2}{3}w_1^*h^2
    \approx\frac{2}{3}w_1^*\epsilon^{2}.
\end{aligned}
\end{equation}
Therefore, we obtain that $w_1^{SAM}\approx w_1^{*}+\frac{2}{3}w_1^*\epsilon^{2}$.

Applying the definition of $W_R^{SAM}$ and $W_R^*$ to this result, we finally reach the conclusion that
\begin{equation}\label{Wsam}
W_R^{SAM}\approx W_R^{*}+\frac{2}{3}W_R^*\epsilon^{2}.
\end{equation}
\end{proof}
\subsection{Proof for Theorem 5}

\begin{proof}
When both methods derives the same robust feature weight $W_R$ (with different perturbation strength $\epsilon_{SAM}$ and $\epsilon_{AT}$), denoting the standard training optimal parameter $W_R^* = (\ln p -\ln (1-p))/2n\eta$, we have \begin{equation}
    W_R=W_{R}^{SAM}\approx W_R^* (1+ \frac{2}{3}\epsilon_{SAM}^{2}) ,
\end{equation}
which is the result of (\ref{Wsam}), and
\begin{equation}
    W_R=W_{R}^{AT}=\frac{\eta}{\eta-\epsilon_{AT}}w_R^*,
\end{equation}
which is the result of (\ref{Wat}) and (\ref{Wst}). Thus, combining two equations, we have
\begin{equation}
    \frac{\eta}{\eta-\epsilon_{AT}}\approx1+ \frac{2}{3}\epsilon_{SAM}^{2}.
\end{equation}
Solving this as an equation of $\eta$, we get relationship\begin{equation}
   \frac{2}{3}\epsilon_{SAM}^{2}\epsilon_{AT}+\epsilon_{AT}\approx\frac{2}{3}\eta\epsilon_{SAM}^{2}.
\end{equation}
By dividing both sides with $\epsilon_{AT}\epsilon_{SAM}^2$, the relation in the theorem can be simply derived. This ends the proof.
\end{proof}

\section{Detailed Experimental Results}
\label{app 2}
\begin{table*}[ht]
Overall experimental results for general robust accuracy are shown in Table~\ref{tab:exp:corruption_cifar10}.
\caption{{General robust accuracy} on \textbf{CIFAR-10C} dataset.}
    \centering
    \begin{tabular}{c|c|cccccccccccc}  
    \toprule
        \multirow{2}{*}{Method}
        &  
        {Natural}
       & \multirow{2}{*}{Brightness} & \multirow{2}{*}{Fog} & \multirow{2}{*}{Frost} & {Gaussian}   
       & {Impulse}
       & \multirow{2}{*}{Jpeg} & {Shot} & \multirow{2}{*}{Snow} & Speckle
       \\
       
       & Accuracy &  & & & Blur & Noise & & Noise &  & Noise\\
       \midrule
        SGD & 94.5 & 63.7 & 26.3 & 30.8 & 19.2 & 40.6 & 32.4 & 26.5 & 41.9 & 32.8 \\
        Adam & 93.9 & 54.8 & 21.9 & 25.4 & 15.9 & 35.7 & 28.1 & 21.5 & 38.0 & 25.1 \\
        \midrule
        SAM ($\rho=0.1$) & 95.4 & 63.1 & 26.2 & 29.2 & 20.6 & 37.0 & 31.7 & 23.0 & 42.7 & 28.3 \\
        SAM ($\rho=0.2$) & \textbf{95.5} & 61.5 & 20.2 & 27.5 & 16.6 & 40.2 & 27.2 & 26.1 & 38.1 & 30.2 \\
        SAM ($\rho=0.3$) & 95.4 & 61.0 & 24.8 & 28.6 & 19.1 & 40.3 & 28.2 & 23.7 & 40.2 & 27.5 \\
        SAM ($\rho=0.4$) & 94.7 & 59.5 & 22.4 & 31.8 & 17.0 & 37.7 & 30.0 & 29.7 & 41.7 & 35.3 \\
        \midrule
        ASAM ($\rho=0.3$) &95.4& \textbf{68.9} & \textbf{28.7}&\textbf{34.0}&18.2&\textbf{41.9}&\textbf{34.0}&25.9&\textbf{45.9}&32.7\\
        ESAM ($\rho=0.3$) &95.2& 64.3 & 28.1&32.9&\textbf{20.9}&39.3&32.8&\textbf{30.0}&44.2&\textbf{34.1}\\
        \midrule
        AT ($\ell_\infty$-$\epsilon=\frac{8}{255}$) & 84.5 & 16.1 & 12.5 & 11.6 & 15.5 & 18.2 & 16.7 & 17.3 & 16.2 & 16.9 \\
        \midrule
        AT ($\ell_2$-$\epsilon=\frac{128}{255}$) & 89.2 & 25.8 & 14.6 & 18.7 & 21.5 & 25.3 & 25.4 & 26.3 & 25.2 & 25.4 \\
        \bottomrule

    \end{tabular}
    \label{tab:exp:corruption_cifar10}
\end{table*}

\end{document}